%% file: manuscript.tex
\documentclass[tikz,manuscript,screen,nonacm]{acmart}

\usepackage{multirow}
\usepackage{graphicx}
\usepackage{tikz}
\usepackage{forest}
\usepackage{arydshln} 
\usepackage{dashrule}
\usepackage[normalem]{ulem}
\usepackage{tikz}
\usepackage{xcolor}
\definecolor{myred}{RGB}{139,0,0}
\definecolor{babyblue}{RGB}{141,192,236}
\definecolor{myorange}{RGB}{238,161,117}
\definecolor{myyellow}{RGB}{251,236,142}
\usetikzlibrary{trees,positioning,shapes,shadows,arrows.meta}

\newcommand\scalemath[2]{\scalebox{#1}{\mbox{\ensuremath{\displaystyle #2}}}}

\AtBeginDocument{%
  }

\setcopyright{acmlicensed}
\copyrightyear{2024}
\acmYear{2024}
\acmDOI{XXXXXXX.XXXXXXX}

\acmJournal{CSUR}
\acmVolume{37} 
\acmNumber{4}
\acmArticle{111}
\acmMonth{7}
\begin{document}

\title{What Makes a Good Story and How Can We Measure It? A Comprehensive Survey of Story Evaluation}


\author{Dingyi Yang}
\email{yangdingyi@ruc.edu.cn}
\author{Qin Jin}
\authornotemark[1]
\email{qjin@ruc.edu.cn}
\affiliation{%
  \institution{Renmin University of China}
  \streetaddress{No.59 Zhongguancun Street}
  \city{Haidian District}
  \country{Beijing, China}
}

\renewcommand{\shortauthors}{Yang and Jin.}

\begin{abstract}
  With the development of artificial intelligence, particularly the success of Large Language Models (LLMs), the quantity and quality of automatically generated stories have significantly increased. This has led to the need for automatic story evaluation to assess the generative capabilities of computing systems and analyze the quality of both automatic-generated and human-written stories. Evaluating a story can be more challenging than other generation evaluation tasks. While tasks like machine translation primarily focus on assessing the aspects of fluency and accuracy, story evaluation demands complex additional measures such as overall coherence, character development, interestingness, etc. This requires a thorough review of relevant research.
  In this survey, we first summarize existing storytelling tasks, including text-to-text, visual-to-text, and text-to-visual. We highlight their evaluation challenges, identify various human criteria to measure stories, and present existing benchmark datasets. Then, we propose a taxonomy to organize evaluation metrics that have been developed or can be adopted for story evaluation. We also provide descriptions of these metrics, along with the discussion of their merits and limitations. Later, we discuss the human-AI collaboration for story evaluation and  generation. Finally, we suggest potential future research directions, extending from story evaluation to general evaluations.
\end{abstract}

\ccsdesc[500]{Computing methodologies~Natural language processing}
\ccsdesc[300]{Computing methodologies~Natural language generation}
\ccsdesc[300]{General and reference~Evaluation}
\ccsdesc[300]{Human-centered computing~Collaborative and social computing design and evaluation methods}
\ccsdesc[300]{Information systems~Multimedia content creation}

\keywords{Storytelling, Data-to-text Generation, Cross-modal generation, Evaluation criteria, Automatic evaluation metrics, Survey}


\maketitle

\input{sections/1.Intro}
\input{sections/2.Tasks}

\input{sections/3.Aspects}

\input{sections/4.Taxonomy}
\input{sections/5.Traditional}
\input{sections/6.LLM_related}
\input{sections/7.Discussions}
\input{sections/8.Collaborative}
\input{sections/9.Future_work}

\input{sections/10.Conclusion}

\bibliographystyle{ACM-Reference-Format}
\bibliography{sample-base}
\end{document}

%% file: sections/1.Intro.tex
\section{Introduction}
Storytelling plays a significant role in human communication. It is widely used in our daily life for various purposes such as education, entertainment, and marketing \cite{Janice2003education,jenkins2014transmedia,Kang2020advertising}. Numerous studies have proposed methods for automatic storytelling, which generate textual narratives from textual \cite{hill2015goldilocks,mostafazadeh2016rocstories,fan2018wpdataset} or visual inputs \cite{krause2017hierarchical, huang2016vist,li2019video}. Another research direction involves animating textual stories with visual content to create a multi-modal story \cite{chen2019neural,rahman2023make,bugliarello2024storybench}.

To improve the quality of automatically generated stories and bridge the gap with human-generated ones, systematic evaluation is crucial. Since storytelling is a creative and open-ended generation task, it is more reasonable to explore metrics based on human standards, rather than only comparing results with the ground truth text \cite{liu2016not}. This necessitates understanding different aspects that humans value, such as fluency, coherence, interestingness, etc. However, the issue of vague and inconsistent evaluation criteria definitions is a long-standing problem \cite{hu2024llm_criteria, zhou2022deconstructing, howcroft2020twenty} that also exists in story evaluation. In this paper, we analyze various criteria in existing works of story evaluation, summarizing the commonly considered aspects for evaluating stories and their definitions. These criteria can improve the reliability and interpretability of evaluation metrics by accessing their correlation with different human standards.

Traditional lexical-based metrics such as BLEU \cite{papineni2002bleu} and ROUGE \cite{lin2004rouge} are widely used, but they often fail to assess semantic aspects \cite{freitag2020bleu_guilty} and show a low correlation with human judgments \cite{sulem2018bleu_not_suitable}. More recent metrics that utilize neural embeddings or generation probabilities, such as BERTScore and \cite{zhang2019bertscore} BARTScore \cite{yuan2021bartscore}, perform better in terms of semantic comprehension, but they are still not particularly effective for evaluating stories. 
Some researchers have proposed metrics \cite{sellam2020bleurt,guan2020union} trained on human evaluation benchmarks, achieving better correlation with human criteria. However, these metrics are still limited by the benchmark itself and the size of the model.

Recently, the development of Large Language Models (LLMs) has led to unprecedented success in understanding and generating text, inspiring research into applying LLMs to evaluation \cite{gao2024llm-based, li2024leveraging, peng2024survey, Chhun2024do}. Compared to traditional methods, LLM-based metrics provide evaluations that are more consistent with human judgment \cite{li2024leveraging}. Additionally, they can provide the reasoning process for the generated score, greatly improving the reliability and interpretability of automatic evaluation scores. This progress also fosters collaborative evaluation \cite{li2023collaborative}, which can leverage both the strength of human and automatic evaluation. Despite the effectiveness of recent methods, there are still under-explored areas in story evaluation, such as personalized evaluation (particularly in terms of subjective aspects like empathy and interestingness), long story evaluation, etc.  These areas still require further research and exploration.

Our survey provides a comprehensive review of story evaluation. We aim to assist researchers in understanding the challenges and progression, as well as identifying potential future directions. The organization of this survey is summarized as follows:
\begin{itemize}
    \item We first summarize the existing story generation tasks and datasets, including text-to-text, visual-to-text, and text-to-visual in Section \ref{sec:tasks}. The evaluation considerations and challenges of various tasks are discussed.
    \item Section \ref{sec:eval_criteria} outlines detailed standards to evaluate stories. To address the issue of vague and inconsistent evaluation criteria, we analyze the commonly considered aspects and their definitions in Section \ref{sec:criteria}, which differ from general NLG tasks \cite{hu2024llm_criteria,howcroft2020twenty}. We then present the existing story evaluation benchmarks and the aspects they cover in Section \ref{sec:benchmark}.
    \item From Section \ref{sec:taxonomy} to \ref{sec:evaluate_metric}, we describe existing metrics using the taxonomy we propose, and explore their correlation with human annotators for story evaluation. Section \ref{sec:taxonomy} introduces a taxonomy to organize existing metrics that have been proposed or can be adopted for story evaluation. We then provide detailed descriptions of traditional metrics in Section \ref{sec:tradition} and LLM-based metrics in Section \ref{sec:llm}. In Section \ref{sec:evaluate_metric}, we specifically discuss the capabilities of different metrics in evaluating stories.
    \item The development of automatic generation and evaluation also encourages human-AI collaborative writing and evaluation. In Section \ref{sec:collaborative}, we discuss the collaborative evaluation, and the methods for measuring collaborative writing systems.
    \item Finally, in Section \ref{sec:future}, we recommend potential future research directions in story evaluation, which can also be extended to general domain.
\end{itemize}

%% file: sections/2.Tasks.tex
\begin{table}[h!]
\centering
\fontsize{7}{8}
\begin{tabular}{cc}
\toprule
\textbf{Type} & \textbf{Tasks} \\
\midrule
\textbf{Text-to-Text} & \begin{minipage}[b]{0.71\columnwidth}
		\raisebox{-.5\height}{\includegraphics[width=\linewidth]{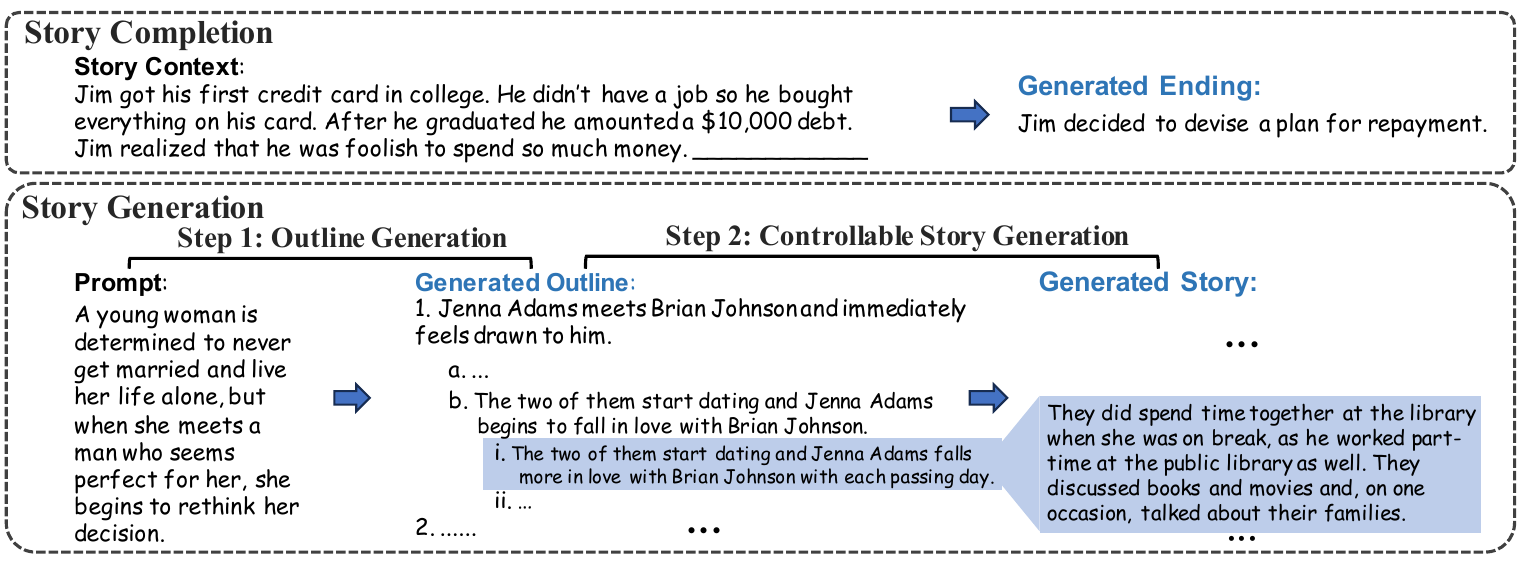}} \end{minipage}\\
\midrule
\textbf{Visual-to-Text} & \begin{minipage}[b]{0.71\columnwidth}
		\raisebox{-.5\height}{\includegraphics[width=\linewidth]{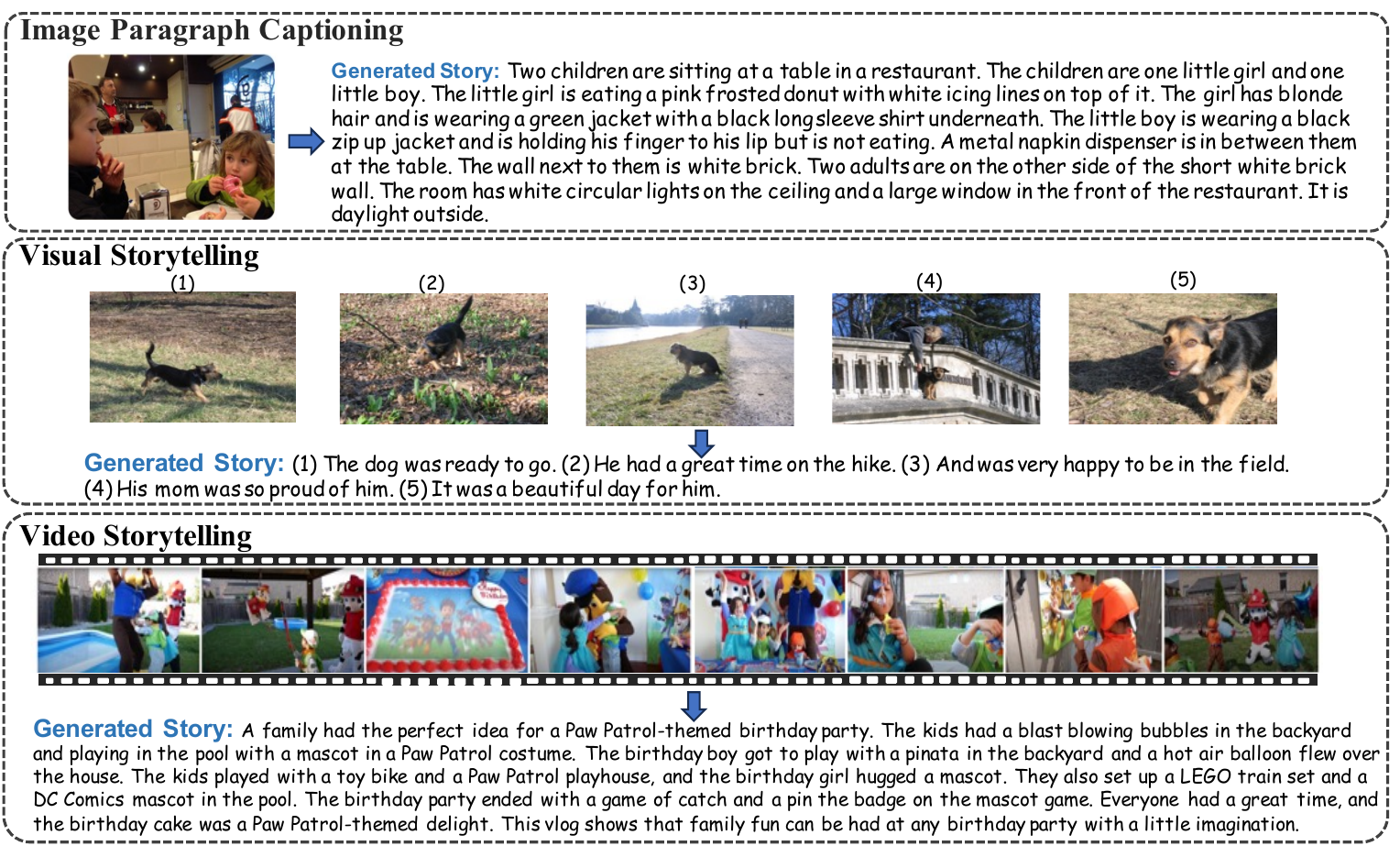}} \end{minipage}\\
  \midrule
\textbf{Text-to-Visual} & \begin{minipage}[b]{0.71\columnwidth}
		\raisebox{-.5\height}{\includegraphics[width=\linewidth]{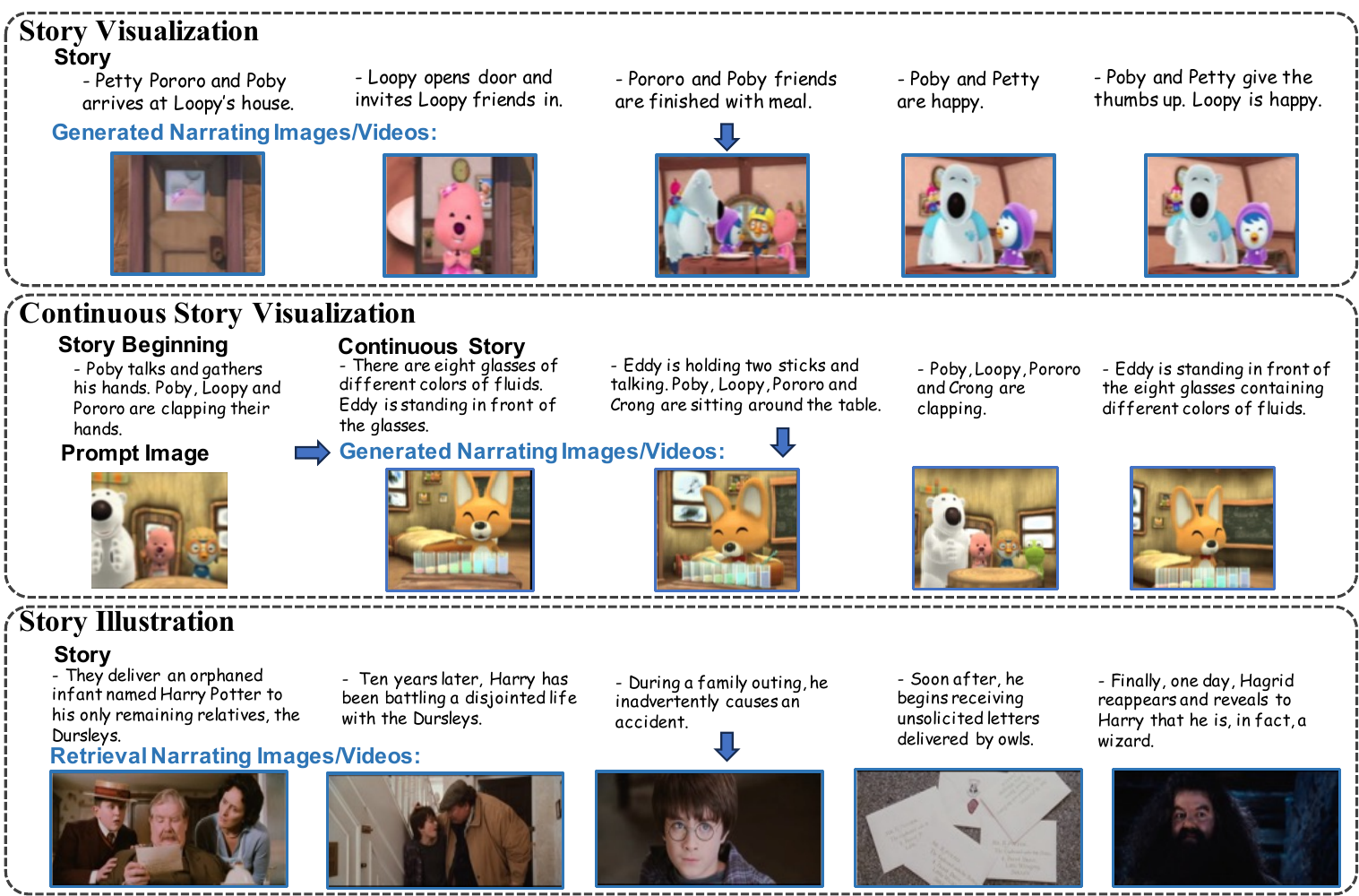}} \end{minipage}\\
\bottomrule
\end{tabular}
\caption{Example of various story generation tasks, including text-to-text (\S \ref{sec:text2text}), visual-to-text (\S \ref{sec:visual2text}), and text-to-visual (\S \ref{sec:text2visual}).}
\vspace{-12pt}
\label{table:task_egs}
\end{table}

\section{Story Generation Tasks} \label{sec:tasks}

In this section, we summarize the existing story generation tasks and discuss the challenges in evaluating them. The tasks are divided into three categories: text-to-text (Section \ref{sec:text2text}), which generates or completes a textual story based on textual inputs; visual-to-text (Section \ref{sec:text2text}), which generates a relevant story based on visual input; text-to-visual (Section \ref{sec:text2text}), which creates visual contents to narrate a textual story. Examples of these various tasks can be found in Table \ref{table:task_egs}, while related datasets are provided in Table \ref{table:tasks}.

\subsection{Text-to-Text} \label{sec:text2text}
Most research on story generation focuses on creating stories based on textual input. Earlier tasks solve the problem of \textbf{Story Completion}, attempting to complete the missing content in the story context. This requires comprehending the given context, then generating missing entities \cite{hill2015goldilocks,hermann2015cnndataset} or spans \cite{ippolito2019unsupervised,donahue2020enabling} within that narrative, or creating a story ending based on the preceding content \cite{mostafazadeh2016rocstories}, as shown in the first line of Table \ref{table:task_egs}. 

More complicated \textbf{Story Generation} tasks aime to create a whole story guided by the following textual controls:
\begin{itemize}
    \item \textit{Title/Topic} \cite{mostafazadeh2016rocstories,Li2013Story} refers to the basic textual input, consisting of a single or a few words that highly summarize the story. For instance, ``going on a date'', ``bank robbery'', etc.
    \item \textit{Prompt/Premise} \cite{fan2018wpdataset,ma2024mops} refers to a more informative input, usually a sentence that captures the main content of the story, including the setting, characters, trajectory, etc. An example is shown in line 2 of Table \ref{table:task_egs}.
    \item \textit{Outline/Plot/Storyline} refers to a list of keywords \cite{yao2019plan}, events \cite{martin2018event}, or sentences \cite{rashkin2020plotmachines} that outline the logical progression of a story, serving as its backbone. For long-form generation, hierarchical outline \cite{Yang2023DOCIL} with a tree-like format can improve both overall and detailed control, as the example shown in line 2 of Table \ref{table:task_egs}.  
    \item \textit{Other Control Signals} include descriptions of the story's main elements, such as characters and scene-settings, which can influence the development of the narrative.
\end{itemize}

Generating an entire story from a basic input such as a topic or a prompt can lead to issues of plot repetition and incoherence, particularly in longer stories \cite{yang2022re3, Yang2023DOCIL}. To address this, some research works \cite{yao2019plan,yang2022re3} have adopted a hierarchical generation process, as shown in Figure \ref{table:task_egs}. The first step, \textbf{Outline Generation}, involves creating a high-level structured outline. This outline can be generated sequentially \cite{rashkin2020plotmachines}, hierarchically \cite{Yang2023DOCIL}, or sampled from a plot graph containing multiple possible event progressions \cite{li2013plot_graph}. The second step develops the outline into a detailed story. This generation process mirrors the way human authors might draft an initial storyline before writing the full story, enhancing the coherence of model-generated stories.

To evaluate the story completion tasks, evaluation metrics are required to judge whether the generated content is fluent, aligned with commonsense, and coherent within the given context. Story generation tasks, on the other hand, are less restrictive. Therefore, the evaluation should consider additional aspects, such as input relevance and interestingness. More aspects and detailed definitions are provided in Section \ref{sec:criteria}.



\subsection{Visual-to-Text}\label{sec:visual2text}
Nowadays, tremendous images and videos can be found on sources like television, the internet, news, and in our daily lives. Generating textual descriptions for these visual resources can help people quickly grasp the main visual content and share it with others \cite{sharma2020image}. Traditional visual captioning tasks generate descriptions that are mundane and may not capture the social relations and emotions within the visual source \cite{huang2016vist}. Recent years, some works \cite{krause2017hierarchical,huang2016vist} attend to craft more interesting and attractive stories for visual inputs, showing both academic and applicable values. For instance, when sharing photos on social media, users may tend to add automatically generated narratives that are coherent and interesting.

\textbf{Image Paragraph Captioning} \cite{krause2017hierarchical} aims to generate a coherent and detailed story for an image. As the example shown in Table \ref{table:task_egs}, it produces a logical description that reflects the detailed content of the image. 
On the other hand, \textbf{Visual storytelling} \cite{huang2016vist} creates a story for an image sequence. Compared to image paragraph captioning, it requires understanding each image and its relationship to each other, constructing a story with a more complex structure and possibly some imaginative content. Video is another common type of visual content. \textbf{Video storytelling} \cite{li2019video} primarily focuses on summarizing sequential events within the video to form a coherent narrative. 

When evaluating a story generated based on visual content, it is crucial to assess both its textual quality and its relevance to the visual input. Specifically, the objects/entities of the story should be characterized based on the depicted visual world, allowing for  reasonable imaginations. For instance, an image showing people dancing could be narrated as ``a group of people are attending a party'', however, ``they are attending a funeral'' would be unreasonable. Additionally, the story plot should be spatially and temporally grounded \cite{halperin2023envisioning} in the visual inputs.


\begin{table}[t]
\caption{Detailed statistics of existing story generation datasets.}
\label{table:tasks} 
\fontsize{5.5}{8}\selectfont
\begin{tabular}{lcccccc}
\toprule
 & \textbf{Dataset} & \textbf{\#Stories} & \textbf{\#Tokens per Story} & \textbf{Annotations} & \textbf{Tasks} &  \textbf{Domain}  \\
 \midrule
\multirow{13}{*}{\textbf{Text-to-Text}} 
& Children's Book \cite{hill2015goldilocks} & 687,343 & 464.7 & Story Context, Query$\rightarrow$Infilling Entity  & Story Completion & Fiction\\
& CNN \cite{hermann2015cnndataset} & 92,579 & 721.9 & Story Context, Query$\rightarrow$Infilling Entity& Story Completion & News \\
 & Story Cloze Test \cite{mostafazadeh2016rocstories} & 3,744 & 48.1 & Story Context$\rightarrow$Ending & Story Completion & Commonsense\\
& RocStories \cite{mostafazadeh2016rocstories} & 98,156 & 88.0 & Title$\rightarrow$Five-Sentence Story & Story Generation & Commonsense\\
& NYTimes \cite{rashkin2020plotmachines,san2008nyt}  & 1,855,658 & - & Title$\rightarrow$Outline\cite{rashkin2020plotmachines}$\rightarrow$Story & Story Generation & News \\
 &  WritingPrompts \cite{fan2018wpdataset} & 303,358 & 735.0 & Prompt$\rightarrow$Outline\cite{rashkin2020plotmachines}$\rightarrow$Story & Story Generation & Real World \\
 & Mystery \cite{ammanabrolu2020bringing} & 532 & 479.4 & Outline$\rightarrow$Story & Story Generation &  Fiction\\
 & Fairy Tales \cite{ammanabrolu2020bringing} & 850 & 543.4 & Outline$\rightarrow$Story & Story Generation &  Fiction\\
 & Hippocorpus \cite{sap2020recollection} & 6,854 & 292.6 & Prompt$\rightarrow$Story & Story Generation &  General \\
 & STORIUM \cite{akoury2020storium} & 5,743 & 19,278 & Prompt, Structural Descriptions$\rightarrow$Story & Story Generation & Fiction \\
 & TVSTORYGEN \cite{chen2021tvstorygen} & 29,013 & 1868.7 & Prompt, Character Descriptions$\rightarrow$Story & Story Generation &  TV Show \\
 & LOT \cite{guan2022lot}  & 2,427 & 128.0 & Title$\rightarrow$Outline$\rightarrow$Story & Story Generation & Fiction \\
 & GPT-BOOKSUM \cite{wang2023improving_pacing}  &  30,047 & 5,363 & Hierarchical outline$\rightarrow$Story & Story/Plot Generation & Fiction \\
 \midrule
\multirow{7}{*}{\textbf{Visual-to-Text}} &  Image Paragraph \cite{krause2017hierarchical} & 19,561 &  67.5 & Image$\rightarrow$Story & Image Paragraph Captioning & Real World\\
& Travel Blogs \cite{park2015expressing} & 11,863 & 222.3  & Image$\rightarrow$Story & Visual Storytelling & Real World\\
& VIST \cite{huang2016vist}  & 50,200 & 57.6 & Image Sequence$\rightarrow$Story & Visual Storytelling & Real World\\
& AESOP \cite{ravi2021aesop} & 7,015 & 26.6 & Image Sequence$\rightarrow$Story  & Visual Storytelling & Real World\\
 & Video Storytelling \cite{li2019video} & 105 & 162.2 & Video$\rightarrow$Story & Video Storytelling & Real World\\
 & VWP \cite{hong2023vwp} & 13,213 & 83.7 & Image Sequence$\rightarrow$Story & Visual Storytelling & Movie \\
 & Album Storytelling \cite{ning2023album} &30 & - & Image Sequence$\rightarrow$Story & Visual Storytelling & Real World\\
 
  \midrule
\multirow{9}{*}{\textbf{Text-to-Visual}} & MUGEN \cite{ hayes2022mugen} & 375,368 & 52.5 & Story$\rightarrow$Video & Story Visualization & Game\\
& PororoSV \cite{li2019storygan} & 15,336 & 69.2 & Story$\rightarrow$Image Sequence & (Continuous) Story Visualization & Cartoon\\
&FlintstonesSV \cite{gupta2018imagine} & 24,512 & 83.1  & Story$\rightarrow$Image Sequence & (Continuous) Story Visualization & Cartoon\\
& DiDeMoSV \cite{maharana2022storydall} & 17,635 & 22.3 & Story$\rightarrow$Image Sequence & (Continuous) Story Visualization & Real World\\
& StorySalon \cite{liu2023intelligent} & 10,366 & 298.1 & Story$\rightarrow$Image Sequence & (Continuous) Story Visualization & Animation \\
& MovieNet-TeViS \cite{gu2023tevis} & 10,000 & 21.7 & Story$\rightarrow$Image Sequence & Story Illustration & Movie \\
& CMD \cite{bain2020condensed} & 3,606 & 136.6& Story$\rightarrow$Video Clip Sequence & Story Illustration & Movie \\
& CVSV \cite{lu2023show} & 84,569 & 534.7 & Story$\rightarrow$Video Clip Sequence & Story Illustration & Movie \\
& StoryBench \cite{bugliarello2024storybench} & 8,900& 37.6 & Story$\rightarrow$Video Segments & (Continuous) Story Visualization & Real World\\
\bottomrule
\end{tabular}
\end{table}

\subsection{Text-to-Visual}\label{sec:text2visual}
With the development of generative models \cite{rombach2021highresolution, kingma2013auto, goodfellow2014generative} capable of generating high-quality images or videos conditioned on textual inputs, there has been an increase in explorations of \textbf{Story Visualization}  \cite{li2019storygan, gupta2018imagine}. This process involves creating visual content to narrate a textual story, making it much more engaging. However, models trained for story visualization may face the challenge of being limited by the characters, backgrounds, and events in the dataset, finding it hard to generate unseen visual content. To address this, \citet{maharana2022storydall} and \citet{bugliarello2024storybench} propose to explore the task of \textbf{Continuous Story Visualization}, which generates visual scenes continuing with a prompt image or video to illustrate the textual story. These works study how the visual scene may change over time to reflect the continuing narrative. Such tasks are less limited by the training dataset, making them more closely aligned with real-world applications. \citet{zang2024let} combine the tasks of visual storytelling and continuous visual storytelling. They first create a textual story based on the events depicted in the input images. Since this story might be incomplete, they forecast further story developments, which contain both textual descriptions and accompanying visual scenes.

To evaluate the visual outputs in story visualization, it is important to assess their quality and their relevance to the input text. Furthermore, they should maintain visual consistency across dynamic visual scenes. For continuous story visualization, the generated scenes, which are constrained by the visual prompt, can be directly compared to the ground truth \cite{bugliarello2024storybench}. Additional evaluation considerations are the same as those for story visualization.

Another direction of the text-to-visual task is \textbf{Story Illustration}, which focuses on retrieving existing images \cite{chen2019neural, gu2023tevis} or videos \cite{lu2023show} to illustrate the textual story. Such tasks overcome the difficulty of generating high-quality visual content, however, the retrieved visuals would be limited by the available visual gallery. Thus, such tasks are usually explored in the domain of movie or TV shows, aiming to create a multi-modal storyboard for a long video.

The evaluation of story illustration should also access the multi-modal relevance and visual consistency. Specifically, such tasks will evaluate retrieval performance by calculating Recall@k \cite{ravi2018show}, measuring the percentage of sentences in the story whose ground truth is in the top-K of retrieved results. Other common retrieval metrics \cite{chen2019neural} include median rank (MedR), mean rank (MeanR), and mean average precision (MAP). \citet{lu2023show} apply the Average k-th Order Precision (AOP-k) \cite{xiong2022transcript} to measure how many sub-sequences are reconstructed in the retrieval visual sequence.

%% file: sections/3.Aspects.tex

\section{Story Evaluation Criteria and Benchmark Datasets} \label{sec:eval_criteria}

To address story evaluation, we must first identify and clearly define various aspects that human evaluation should consider. The issue of vague and inconsistent evaluation criteria is a long-standing problem and has been explored in the general evaluation domain \cite{hu2024llm_criteria, zhou2022deconstructing, howcroft2020twenty}.
Compared to other tasks like machine translation, story generation introduces a more diverse set of evaluation aspects. The definitions of the same aspects may also differ. For instance, characters can play an important role in narratives -- an aspect overlooked in other tasks. This introduces the specific aspect of character development. Moreover, if characters are crucial to plot progression, they will play a significant role in the evaluation of coherence as well. 

In Section \ref{sec:criteria}, we summarize the story evaluation criteria, including the considered aspects and their definitions. We then present existing story evaluation benchmarks in Section \ref{sec:benchmark}.

\subsection{Story Evaluation Criteria} \label{sec:criteria}
We analyze various aspects proposed in  existing story evaluation research \cite{wang2023perse,maimon-tsarfaty-2023-cohesentia,xie2023can,chhun2022hanna,guan2021openmeva,guan2020union,guan2020knowledge}, as well as evaluation criteria considered in existing story generation works. We summarize commonly used aspects, and integrate their definitions as shown in Table \ref{table:criteria}. These aspects are organized in a hierarchical structure, where some are sub-aspects of others. Among these commonly explored aspects, the following are more subjective and could be influenced by personal preference \cite{wang2023perse}: character development, interestingness, empathy, and surprise. For user-oriented scenarios like recommendation systems and search applications \cite{Abhinandan2007news, Croft2001Relevance}, personal preference should be considered in the evaluation.

Except for the commonly considered aspects in Table \ref{table:criteria}, other factors that might be considered under specific conditions include:

\noindent \textbf{- Style.} Whether the writing style, formality, or tone remains consistent throughout the whole story \cite{roemmele2017evaluating}. This is achieved by using proper words, sentences, quotes, terms, and so on. In previous works on stylized story generation \cite{prabhumoye2019my,mou2020stylized,kong2021stylized}, such ``style'' can be simplified as a specific emotion \cite{brahman2020modeling} or sentiment 
\cite{peng2018towards, luo2019learning}, where the performance is evaluated by style classification accuracy. 

\noindent \textbf{- Controllable Accuracy.} In the task of controllable storytelling, guided by elements such as scene descriptions \cite{akoury2020storium} and character relationships \cite{vijjini2022towards}, it is necessary to evaluate the controllable accuracy, which means whether the story content correctly reflects the controllable signals.

\noindent \textbf{- Toxicity.} Whether the story includes some rude, unreasonable, or disrespectful components. This aspect is quite important for children stories \cite{bhandari2023trustworthiness}.

\noindent \textbf{- Naturalness/Human-like.}  Whether the model-generated story is likely to be written by a human \cite{pillutla2021mauve}. 

\noindent \textbf{- Non-Hallucination.} Whether the generated story contains unreasonable information that cannot be supported by the source input \cite{huang2023hallucination}. As story generation is a creative task, this aspect is randomly considered. It might be considered in visual storytelling, referring to unreasonable imagination. For instance, in the visual story shown in line 5 of Table \ref{table:task_egs}, it is reasonable to state that ``his mom was proud of him'' when observing the owner playing with her dog. However, ``his mom played frisbee with him'' would be considered a hallucination, as there are no visual elements related to a frisbee. Imagination is acceptable, but it has to be reasonable.

\noindent \textbf{- Visual Quality.} Specifically for text-to-visual story generation tasks, it is required to measure the visual quality and visual consistency of the generated visual content, as detailed in Section \ref{sec:visual_quality}. 

\begin{table}[h!]
\caption{Common aspects used for evaluating story quality and their definitions.}
\label{table:criteria} 
\fontsize{7.6}{11}\selectfont
\begin{tabular}{lll}
\toprule
\multicolumn{2}{c}{\textbf{Aspect}} & \multicolumn{1}{c}{\textbf{Definition}} \\
\midrule
\multicolumn{2}{l}{\textbf{Relevance}} & \begin{tabular}[c]{@{}l@{}}Whether the story is relevant to and reasonably reflects the source input. \end{tabular} \\
\midrule
\multicolumn{2}{l}{\textbf{Diversity}} & Whether the stories generated by one model have many variations. \\
\midrule
\multicolumn{2}{l}{\textbf{Fluency}} & \begin{tabular}[c]{@{}l@{}}Whether the individual sentences within a story are of high quality. They should be grammatically correct, \\ free of typos, non-repetitive, and in line with common language usage.\end{tabular} \\
\hdashline[1.5pt/2.5pt]
 &\textbf{Grammaticality} & \begin{tabular}[c]{@{}l@{}}Whether the individual sentences are grammatically correct without lexical or syntax errors.  Note it focuses \\on syntax only, not semantics.\end{tabular} \\
 & \textbf{Non-redundancy} & Whether the individual sentences are free of redundant elements, such as repetition, over-specificity, etc. \\
\midrule
\multicolumn{2}{l}{\textbf{Coherence}} & \begin{tabular}[c]{@{}l@{}}Whether all sentences and plots are well structured, with the context organized and connected logically. \\ Evaluating coherence usually ignores grammar or spelling errors.\end{tabular} \\
\hdashline[1.5pt/2.5pt]
 & \textbf{Cohesion} & \begin{tabular}[c]{@{}l@{}}Whether the sentences in a story are formally connected. They can be connected by either referential links \\(co-reference,  bridging anaphora) or by semantic connectors \cite{maimon-tsarfaty-2023-cohesentia}.\end{tabular} \\
 & \textbf{Consistency} &  Whether the sentences are logically aligned with the preceding story.\\
 & \textbf{Implicit Relevance} & Whether a story follows the same topic from beginning to end. \\
 \midrule
\multicolumn{2}{l}{\textbf{Completeness}} & \begin{tabular}[c]{@{}l@{}} Whether the story covers all its underlying concepts, theories, and historical context. \end{tabular} \\
\hdashline[1.5pt/2.5pt]
 & \textbf{Ending} & Whether the story has a clear and rational ending.\\
 \midrule
\multicolumn{2}{l}{\textbf{Clarity}} & \begin{tabular}[c]{@{}l@{}}Whether the story is clear and easy to understand, with no confusing or ambiguous elements.\end{tabular} \\
\midrule
\multicolumn{2}{l}{\textbf{Commonsense}} & \begin{tabular}[c]{@{}l@{}} Whether the story adheres to commonsense knowledge, such as physical entities and social interactions.  \end{tabular} \\
\midrule
\multicolumn{2}{l}{\textbf{Informativeness/Complexity}} & \begin{tabular}[c]{@{}l@{}} Whether the story contains rich and detailed information to support its progression and world-building. \end{tabular} \\

\midrule
\multicolumn{2}{l}{\textbf{Character Development}} & \begin{tabular}[c]{@{}l@{}}Whether the story features well-developed and engaging characters that are believable, relatable, and \\ contribute to the overall narrative or theme.\end{tabular} \\
\midrule
\multicolumn{2}{l}{\textbf{Interestingness/Engagement}} & \begin{tabular}[c]{@{}l@{}}Whether the story is highly enjoyable or entertaining to read, with rich details and descriptions that engage \\ the readers’  senses and imagination.\end{tabular} \\
\midrule
\multicolumn{2}{l}{\textbf{Empathy}} & \begin{tabular}[c]{@{}l@{}} Whether the story arouses the readers’ emotional experience, passion, and empathy. \end{tabular} \\
\midrule
\multicolumn{2}{l}{\textbf{Surprise}} & \begin{tabular}[c]{@{}l@{}}Whether the story creates suspense and surprise, especially in mystery fictions. \end{tabular} \\
\bottomrule

\end{tabular}
\end{table}

\subsection{Story Evaluation Benchmark Datasets} \label{sec:benchmark}
\begin{figure}[h!]
    \centering  \includegraphics[width=0.83\linewidth]{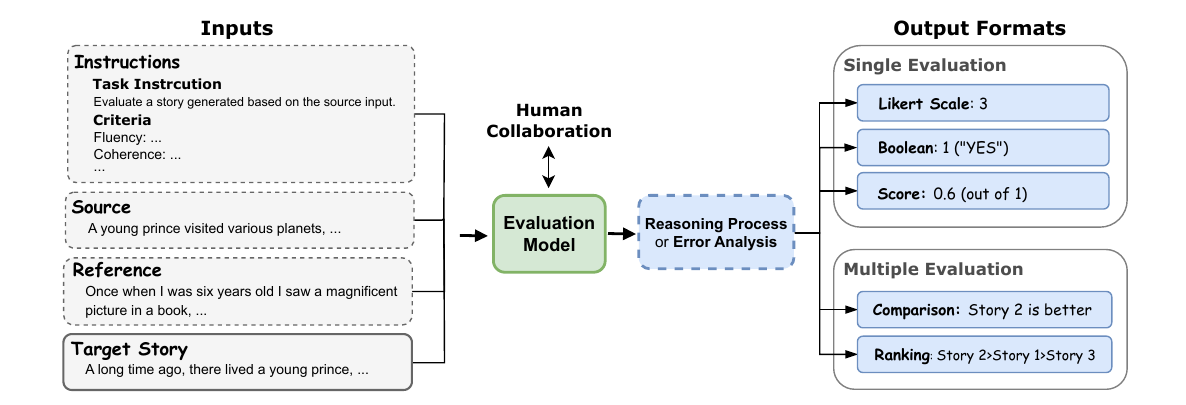}
    \caption{General Framework of Story Evaluation, which shows the evaluation inputs and output formats (Section \ref{sec:format}). All the dashed boxes are optional input or output.}
    \label{fig:eval_frame}
\end{figure}

We present detailed statistics of existing story evaluation benchmark datasets in Table \ref{table:eval_dataset}. As displayed in Figure \ref{fig:eval_frame}, annotators are required to evaluate a single story or compare multiple stories, with or without the source input and reference text. Some datasets measure the overall quality of the story, while others evaluate a specific aspect \footnote{We convert a few original aspects into our summarized aspects (\S \ref{sec:criteria}). For the original aspect names and definitions, please refer to the related papers.}. Among these datasets, OpenMEVA and HANNA are the most cited. OpenMEVA \cite{guan2021openmeva} propose evaluation on overall quality, while HANNA \cite{chhun2022hanna} annotates evaluation scores on multiple aspects of a story. Notably, COHESENTIA \cite{maimon-tsarfaty-2023-cohesentia} firstly introduces a benchmark for the vague aspect of coherence, evaluating both global and local coherence (scoring sentence by sentence). PERSE \cite{wang2023perse} firstly focuses on personalized story evaluation, considering the readers' personal preferences.

\begin{table}[h!]
\caption{The statistics of human-annotated story evaluation benchmark datasets. ``Format'' 
and ``Reasoning'' refers to the evaluation format (Sec. \ref{sec:format}) and whether the reasoning process is annotated; ``Criteria'' denotes whether each sample considers the Overall Quality or a Single Aspect, while ``Aspects'' refers to the considered aspects; ``\#Stories'' and ``\#Samples'' refers to the correlated data size. For the abbreviation of each aspect, REL: relevance, FLU: fluency, COH: coherence, END: ending, CLA: clarity, COMM: commonsense, INF: informativeness,  CHA: character development, INT: interestingness, ADAP: adaptability (whether a plot could guide the generation of a interesting story \cite{wang2023perse}), EMP: empathy, SUR: surprise, STY: style.}
\label{table:eval_dataset} 
\fontsize{6}{7.5}\selectfont
\begin{tabular}{cccccccc}
\toprule
\textbf{Dataset} & \textbf{Story Type} & \textbf{Format} & \textbf{Reasoning} & \textbf{Criteria} & \textbf{Aspects} & \textbf{\#Stories} & \textbf{\#Samples}\\
\midrule
OpenMEVA \cite{guan2021openmeva}& Model-Generated Story & Likert Scale (1-5)& No  & Overall & REL, FLU, COH, COMM & 2,000 & 2,000 \\
HANNA \cite{chhun2022hanna} & Model-Generated  Story& Likert Scale (1-5)& No & Single & REL, COH, EMP, SUR, INT, INF  & 1,056  & 19,008 \\
VHED \cite{hsu2022learning} & Model-Generated Visual Story & Comparison & No& Overall  & FLU, COH, CLA, REL & 4,500 & 13,875\\
StoryER-Rank \cite{chen2023storyer} & Human-Written Story & Comparison & No & Overall  & - & 63,929 & 116,971\\
StoryER-Scale \cite{chen2023storyer} & Human-Written Story &  Likert Scale (1-5) & Yes & Single  & COH, END, STY, CHA, EMP  & 12,669 & 45,948\\
Per-MPST \cite{wang2023perse} &  Human-Written Movie Plot & Comparison & Yes & Overall & - & 981 & 69,947 \\
Per-DOC \cite{wang2023perse} &  Human-Written Novel Plot & Likert Scale (1-5) & Yes & Single  & INT, ADAP, SUR, CHA, END & 596  & 8,922 \\
Xie \cite{xie2023can} &  Model-generated Story & Likert Scale (1-5) & No & Single  & REL, FLU, COH, COMM, INT & 200  & 1,000 \\
COHESENTIA \cite{maimon-tsarfaty-2023-cohesentia} & Model-generated Story & Likert Scale (1-5) & Yes & Single & COH & 500 & 500 \\
\bottomrule
\end{tabular}
\end{table} 

\vspace{-0.5cm}

\vspace{-0.3cm}


%% file: sections/4.Taxonomy.tex
\section{Taxonomy of Evaluation Metrics} \label{sec:taxonomy}

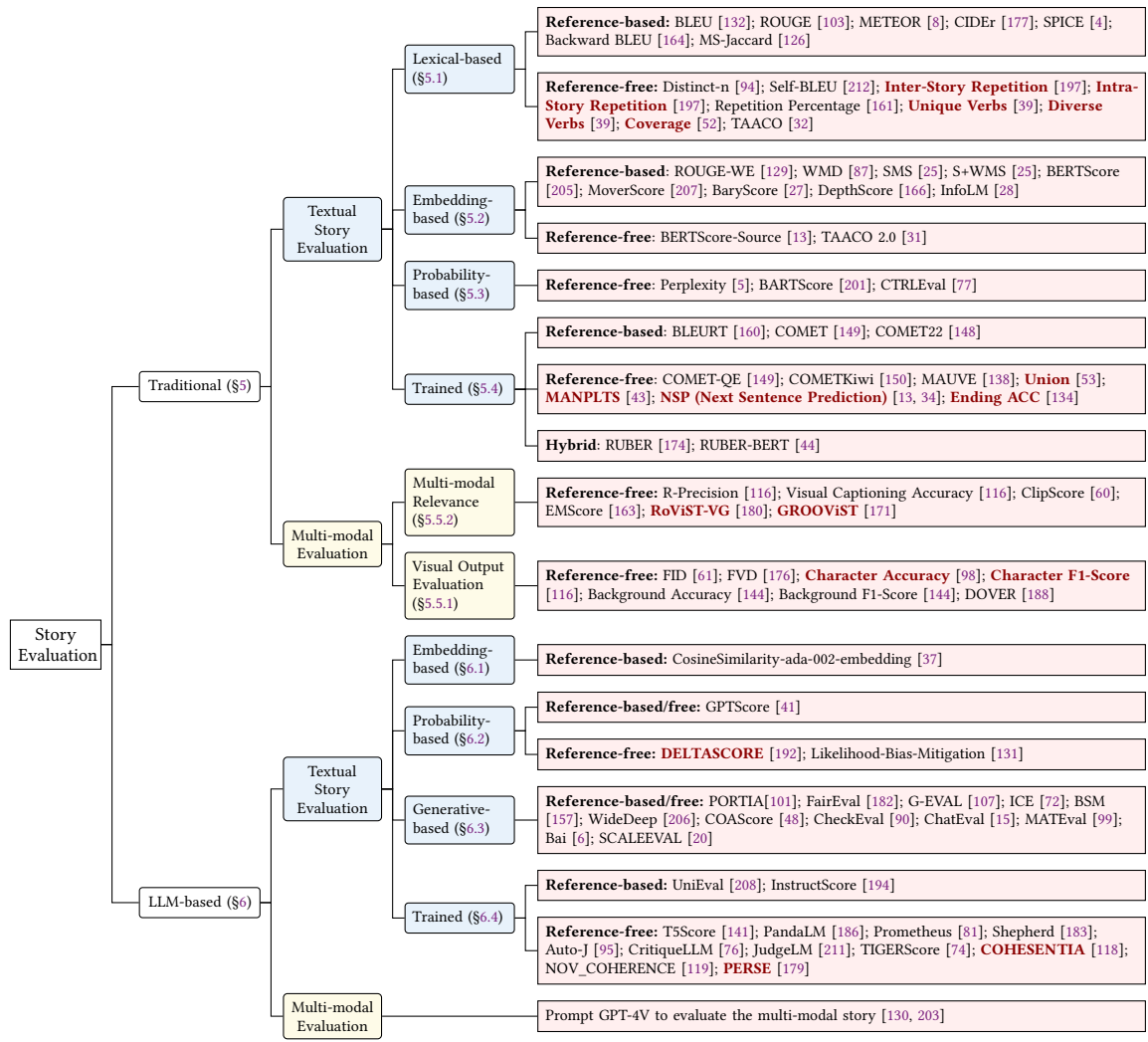
\begin{figure}[t]
    \centering
    
\tikzset{
    basic/.style  = {draw, text width=1cm, align=center, font=\sffamilyfont,font=\scriptsize, rectangle},
    root/.style   = {basic, rounded corners=1.5pt, thin, align=center},
    onode/.style = {basic, thin, rounded corners=1.5pt, align=center,text width=3cm,},
    tnode/.style = {basic, thin, fill=pink!25, text width=25em, align=left},
    tnode_short/.style = {basic, thin, align=left, fill=pink!50, text width=20.5em, align=left},
    xnode/.style = {basic, thin, rounded corners=1.3pt, fill=babyblue!20, align=left, text width=1.25cm,},
    mm_xnode/.style = {basic, thin, rounded corners=1.3pt, fill=myyellow!20,  align=left, text width=1.25cm,},
    xnode_long/.style = {basic, thin, rounded corners=1.2pt, align=center, text width=1.4cm,},
    textnode/.style = {basic, thin, rounded corners=1.5pt, fill=babyblue!20, align=center, text width=1.1cm,},
    mmnode/.style = {basic, thin, rounded corners=1.5pt, fill=myyellow!20, align=center, text width=1.1cm,},
    wnode/.style = {basic, thin, align=left, fill=pink!10!blue!80!red!10, text width=6.5em},
    edge from parent/.style={draw=black, edge from parent fork right}

}

\begin{forest} for tree={
    grow=east,
    growth parent anchor=west,
    parent anchor=east,
    child anchor=west,
    calign=center,
    anchor=center,
    if n children=0{tier=last}{},
    edge path={
    \noexpand\path [draw, \forestoption{edge}] (!u.parent anchor) -- +(4pt,0) |- (.child anchor)\forestoption{edge label};
  }
}
[Story \\Evaluation, basic,  l sep=5mm, font=\footnotesize
  [LLM-based (\S \ref{sec:llm}), xnode_long,  l sep=3mm,
    [Multi-modal Evaluation, mmnode,  l sep=3mm,
        [Prompt GPT-4V to evaluate the multi-modal story \cite{ning2023album,zhang2023mm-narrator},tnode]
    ]
    [Textual Story Evaluation, textnode,  l sep=3mm,
      [Trained (\S \ref{sec:llm_trained}), xnode,  l sep=3mm,
        [\textbf{Reference-free:}  T5Score \cite{qin2022t5score}; PandaLM \cite{wang2023pandalm}; Prometheus \cite{kim2023prometheus}; Shepherd \cite{wang2023shepherd}; Auto-J \cite{li2023autoj}; CritiqueLLM \cite{ke2023critiquellm}; JudgeLM \cite{zhu2023judgelm}; TIGERScore \cite{jiang2023tigerscore}; \textcolor{myred}{\bf COHESENTIA} \cite{maimon-tsarfaty-2023-cohesentia}; NOV\_COHERENCE \cite{maimon2023novel_coh}; \textcolor{myred}{\bf PERSE} \cite{wang2023perse}, tnode]
        [\textbf{Reference-based:} UniEval \cite{zhong2022unieval}; InstructScore \cite{xu2023instructscore}, tnode]
      ]
      [Generative-based (\S \ref{sec:llm_gen}), xnode,  l sep=3mm,
        [\textbf{Reference-based/free:} PORTIA\cite{li2023split}; FairEval \cite{wang2023faireval}; G-EVAL \cite{liu2023gpteval}; ICE \cite{jain2023multi}; BSM \cite{saha2023branch}; WideDeep \cite{zhang2023wider}; COAScore \cite{gong2023coascore}; CheckEval \cite{lee2024checkeval}; ChatEval \cite{chan2023chateval}; MATEval \cite{li2024mateval}; Bai \cite{bai2024benchmarking}; SCALEEVAL \cite{chern2024scaleeval}, tnode]
      ]
      [Probability-based (\S \ref{sec:llm_pro}), xnode,  l sep=3mm,
        [\textbf{Reference-free:} \textcolor{myred}{\bf DELTASCORE} \cite{xie2023deltascore}; Likelihood-Bias-Mitigation \cite{ohi2024likelihood} , tnode]
        [\textbf{Reference-based/free:} GPTScore \cite{fu2023gptscore}, tnode]
      ]
      [Embedding-based (\S \ref{sec:llm_embedding}), xnode,  l sep=3mm,
        [\textbf{Reference-based:} CosineSimilarity-ada-002-embedding \cite{es2023ragas}, tnode]
      ]
    ]
  ]
  [Traditional (\S \ref{sec:tradition}), xnode_long, l sep=3mm,
    [Multi-modal Evaluation, mmnode,  l sep=3mm,
        [Visual Output \\ Evaluation (\S \ref{sec:multi_modal_rel}), mm_xnode,  l sep=3mm,
            [\textbf{Reference-free:}  {FID} \cite{heusel2017gans}; {FVD} \cite{unterthiner2019fvd}; \textcolor{myred}{\bf Character Accuracy }\cite{li2019storygan}; \textcolor{myred}{\bf Character F1-Score} \cite{maharana2021improving}; Background Accuracy \cite{rahman2023make}; Background F1-Score \cite{rahman2023make}; DOVER \cite{wu2022disentangling}, tnode]
        ]
        [Multi-modal Relevance (\S \ref{sec:visual_quality}), mm_xnode,  l sep=3mm,
            [\textbf{Reference-free:} R-Precision \cite{maharana2021improving}; Visual Captioning Accuracy \cite{maharana2021improving}; ClipScore \cite{hessel2021clipscore}; EMScore \cite{emscore}; \textcolor{myred}{\bf RoViST-VG} \cite{wang2022rovist}; \textcolor{myred}{\bf GROOViST} \cite{surikuchi2023groovist}, tnode]
        ]
    ]
    [Textual Story Evaluation, textnode,  l sep=3mm,
        [Trained (\S \ref{sec:trained}), xnode,  l sep=3mm,
        [\textbf{Hybrid}: RUBER \cite{tao2018ruber}; RUBER-BERT \cite{tao2018ruber_bert} , tnode]
        [\textbf{Reference-free}: COMET-QE \cite{rei2020comet}; COMETKiwi \cite{rei2022cometkiwi}; MAUVE \cite{pillutla2021mauve}; \textcolor{myred}{\bf Union} \cite{guan2020union}; \textcolor{myred}{\bf MANPLTS} \cite{ghazarian2021MANPLTS}; \textcolor{myred}{\bf NSP (Next Sentence Prediction)} \cite{callan2023interesting, devlin2018bert}; \textcolor{myred}{\bf Ending ACC} \cite{park2023longstory}, tnode]
        [\textbf{Reference-based}: BLEURT \cite{sellam2020bleurt}; COMET \cite{rei2020comet}; COMET22 \cite{rei2022comet} , tnode]
      ]
      [Probability-based (\S \ref{sec:probability}), xnode,  l sep=3mm,
        [\textbf{Reference-free}: Perplexity \cite{Bahl_Jelinek_Mercer_1983}; 
 BARTScore \cite{yuan2021bartscore}; CTRLEval \cite{ke2022ctrleval}, tnode]
    ]
      [Embedding-based (\S \ref{sec:embedding}), xnode,  l sep=3mm,
        [\textbf{Reference-free}: BERTScore-Source \cite{callan2023interesting}; TAACO 2.0 \cite{crossley2019tool}, tnode]
        [\textbf{Reference-based}: ROUGE-WE \cite{ng2015rouge_we}; WMD \cite{kusner2015word}; SMS \cite{clark2019sentence}; S+WMS \cite{clark2019sentence}; BERTScore \cite{zhang2019bertscore}; MoverScore \cite{zhao2019moverscore}; BaryScore \cite{colombo-etal-2021-automatic}; DepthScore \cite{depth_score}; InfoLM \cite{infolm_aaai2022}, tnode]
      ]
      [Lexical-based (\S \ref{sec:lexical}), xnode,  l sep=3mm,
        [\textbf{Reference-free:} Distinct-n \cite{li2015diversity}; Self-BLEU \cite{zhu2018self_bleu}; \textcolor{myred}{\bf Inter-Story Repetition} \cite{yao2019plan}; \textcolor{myred}{\bf Intra-Story Repetition} \cite{yao2019plan}; Repetition Percentage \cite{shao2019long}; \textcolor{myred}{\bf Unique Verbs} \cite{fan2019strategies}; \textcolor{myred}{\bf Diverse Verbs} \cite{fan2019strategies}; \textcolor{myred}{\bf Coverage} \cite{guan2020knowledge}; TAACO \cite{crossley2016tool}, tnode]
        [\textbf{Reference-based:} BLEU \cite{papineni2002bleu}; ROUGE \cite{lin2004rouge}; METEOR \cite{banerjee2005meteor}; CIDEr \cite{vedantam2015cider}; SPICE \cite{Anderson2016spice}; Backward BLEU \cite{shi2018bbl}; MS-Jaccard \cite{montahaei2019jointly} , tnode]
      ]
    ]
  ]
]
\end{forest}

    \caption{Taxonomy of evaluation metrics proposed or can be adopted for story evaluation. The metrics that are specifically proposed for story evaluation are \textcolor{myred}{\bf colored}.}
    \label{fig:lit_surv} 
    \vspace{-0.4cm}
\end{figure}

So far, we have discussed current story generation tasks and what characteristics make a good story. As human evaluation can be time-consuming and labor-intensive, in recent years, several automatic metrics have been proposed or can be adopted for story evaluation. This section proposes a taxonomy to organize existing metrics, as illustrated in Figure \ref{fig:lit_surv}. We categorize them broadly into  \textbf{traditional} (Section \ref{sec:tradition}) and \textbf{LLM-based} methods (Section \ref{sec:llm}). LLM-based metrics refer to those based on models with billion-level parameters. 


Considering the task type of story generation,
both text-to-text and visual-to-text tasks require \textbf{textual story evaluation}, measuring the quality of the generated textual story. \textbf{Multi-modal evaluation} is required for cross-modal generation, including metrics to measure multi-modal relevance (for both visual-to-text and text-to-visual tasks) and the quality of visual output (for text-to-visual tasks). 

Based on the implementation methods of text evaluation metrics, we categorize them into five types, as detailed in Section \ref{sec:method_type}. We also classify their output formats, as explained in Section \ref{sec:format}. 
Note that most story-related generation tasks do not have a standard result, with exceptions like story entity completion and continuous story visualization. Therefore, while \textbf{reference-based} metrics (comparing target text to reference text) can be useful, \textbf{reference-free} metrics (evaluating target text based on its overall or multi-aspect quality) may be more appropriate. There are also some \textbf{hybrid} metrics that combine reference-free and reference-based results.

As discussed in our introduction, metrics that measure specific aspects of a story can enhance the reliability and interpretability of automatic evaluation compared to those assessing overall quality. Table \ref{table:all_metrics} provides a comprehensive overview of metrics evaluating various aspects of textual stories.

\begin{figure}[t]
    \centering
    \includegraphics[width=0.98\linewidth]{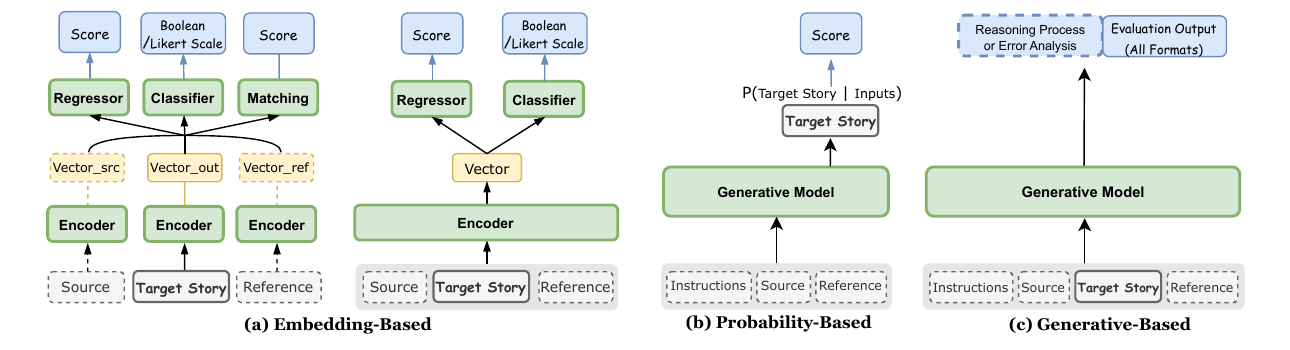}
    \caption{Illustration of different types of neural models applied for automatic evaluation metrics (all the dashed boxes are optional input or output): (a) Embedding-Based Methods, which evaluate based on separately encoded vectors (left) or a jointly encoded vector (right); (b) Probability-Based Methods, which calculate based on the generation probability of the target story; (c) Generative-Based Methods, which directly generate the evaluation results, with or without the reasoning process. These three types of models can be fine-tuned on evaluation benchmarks, referred to as Trained Metrics.}
    \label{fig:eval_model}
\end{figure}

\subsection{Evaluation Methods} \label{sec:method_type}
According to the implementation methods, we categorize the evaluation metrics into the following types :

\noindent{\textbf{- Lexical-Based.}} Such methods process the story as bag-of tokens or n-grams (n continuous tokens). They are usually used to measure the similarity between two stories or to assess the diversity of stories generated by one model.

\noindent{\textbf{- Embedding-Based.}} As shown in the left part of Figure \ref{fig:eval_model} (a), some embedding-based approaches use pre-trained models to encode the source input, target story, and its reference as multiple embeddings, then perform further processing. Specifically, the matching-based model measures the equivalence between the target and reference vector, while the regressor or classifier achieves the result with or without the reference. As displayed in the right part of Figure \ref{fig:eval_model} (a), other embedding-based methods encode all inputs jointly to obtain a vector for further calculation.

\noindent{\textbf{- Probability-Based.}} These types of methods calculate the evaluation score based on the generation probability of the target text through generative models, as illustrated in Figure \ref{fig:eval_model} (b).

\noindent{\textbf{- Generative-Based.}} These methods can also be referred to as prompt-based methods. They provide humans or generative models (Figure \ref{fig:eval_model} (c)) with an evaluation prompt and collect the generated results. As an example shown in Figure \ref{fig:eval_frame}, the input prompt provides the instructions, including the \textit{task instruction} and \textit{specific criteria}. It also provides information about the evaluation sample, including the \textit{source input}, \textit{target story}, and \textit{reference story}.

\noindent{\textbf{- Trained.}} These methods might employ any of the three model types shown in Figure \ref{fig:eval_model}, with further training on evaluation benchmark datasets to improve the evaluating abilities. Some approaches also incorporate training on specially designed tasks using the story generation dataset to improve text comprehension. In the ``Method'' column of Table \ref{table:all_metrics}, we present the model types of the trained metrics, i.e., embedding-based (trained), probability-based (trained), and generative-based (trained).

\subsection{Evaluation Output Format} \label{sec:format}
The output format types of different metrics can be categorized as follows:

\noindent{\textbf{- Score.}} Evaluate a discrete score, usually on a scale of 0-1 or 0-100, such as BLEU, ROUGE, etc. 

\noindent{\textbf{- Likert Scale \cite{Robertson2012Likert}.}} Rate on an integer scale, usually from 1-5, with 1 being the lowest scale and 5 the highest.

\noindent{\textbf{- Boolean.}} Evaluate if one story is good or bad, or whether it meets certain criteria (for example, checking whether it is fluent).

\noindent{\textbf{- Comparison.}} Choose the better of two stories. Previous works \cite{liusie2024comparative} have proposed that it is easier and more robust to compare two outputs than to score them independently.

\noindent{\textbf{- Ranking.}} For  comparison, only two stories are involved, while for ranking, the order of N (N$\geq$2) samples needs to be decided. Specifically, we can get the ranking list through N(N-1) pairwise comparisons or calculate a win-loss ratio within selected comparisons to order the candidates \cite{liusie2024comparative}.

\noindent{\textbf{- Error Analysis.}} 
To ensure the trustworthiness and reliability of evaluation results, some studies \cite{xie2023deltascore} provide a detailed evaluation process for each score, including specific error locations and explanations \footnote{Reasoning process, on the other hand, provides a more coarse-grained explanation of the evaluation result.}. For instance, TIGERScore \cite{jiang2023tigerscore} assigns a penalty score to each identified error and then calculates the discrete score. Alternatively, as suggested in \citet{hu2023decipherpref}, one could simply count the number of errors to categorize the score from low to high scale. For aspects such as commonsense, we can calculate the error percentage as done in \citet{min2023factscore}.


\begin{table}[t]
\caption{Evaluation Metrics proposed (\checkmark) and adopted (*) for evaluating various aspects of generated texts. For each aspect, REL: relevance, DIV: diversity, FLU: fluency, COH: coherence, COM: completeness, CLA: clarity, COMM: commonsense, INF: informativeness, CHA: character development, INT: interestingness, EMP: empathy, SUR: surprise. }
\label{table:all_metrics} 
\fontsize{6}{8.2}\selectfont
\begin{tabular}{ccccccccccccccc}
\toprule
\multicolumn{1}{c}{\textbf{Metric}} & \multicolumn{12}{c}{\textbf{Aspects}} & \multicolumn{1}{c}{\textbf{Method}} & \multicolumn{1}{c}{\textbf{Format}} \\
\textbf{} & \makebox[0.018\textwidth][c]{REL} & \makebox[0.018\textwidth][c]{DIV} & \makebox[0.018\textwidth][c]{FLU} & \makebox[0.018\textwidth][c]{COH} & \makebox[0.018\textwidth][c]{COM} & \makebox[0.018\textwidth][c]{CLA} & \makebox[0.018\textwidth][c]{COMM} & {INF} &  \makebox[0.018\textwidth][c]{CHA} & \makebox[0.018\textwidth][c]{INT} &\makebox[0.018\textwidth][c]{EMP} & \makebox[0.018\textwidth][c]{SUR} &    \\
\midrule
Distinct-n \cite{li2015diversity} &  & \checkmark & & &  &  &  &  &  &  &  & & Lexical-based & Score \\
SELF\_BLEU \cite{zhu2018self_bleu} &  & \checkmark &  &  &  &  &  &  &  &  &  & & Lexical-based & Score \\
Backward BLEU \cite{shi2018bbl} &  & \checkmark &  &  &  &  &  &  &  &  & &  & Lexical-based & Score \\
MS-Jaccard \cite{montahaei2019jointly} &  & \checkmark &  &  &  &  &  &  &  &  &  &  & Lexical-based & Score \\
Inter-Story Repetition \cite{yao2019plan} &  & \checkmark &  &  &  &  &  &  &  &  &  &  & Lexical-based & Score \\
Intra-Story Repetition \cite{yao2019plan} & & & \checkmark &  &  &  &  &  &  &  &  &  & Lexical-based & Score \\
TAACO \cite{crossley2016tool} & & &  & \checkmark &  &  &  &  &  &  &  &  & Lexical-based & Likert \\
TAACO 2.0 \cite{crossley2019tool} & & &  & \checkmark &  &  &  &  &  &  &  &  & Embedding-based & Likert \\
UNION \cite{guan2020union} &  &  & \checkmark & \checkmark & &  & \checkmark  &  &  &  &  &  & Embedding-based (trained) & Boolean \\
MANPLTS \cite{ghazarian2021MANPLTS} &  &  &  & \checkmark & &  &  &  &  &  &  &  & Embedding-based (trained) & Boolean \\
BERTScore-Source \cite{callan2023interesting} & \checkmark &  &  &  &   &  &  &  &  &  &  &  & Embedding-based & Score \\ %
Perplexity \cite{Bahl_Jelinek_Mercer_1983} &  &  & \checkmark & \checkmark &  &  &  &  &  &  &  &  & Probability-based & Score\\
BARTScore \cite{yuan2021bartscore} & \checkmark &  & \checkmark & \checkmark &  &  &  &  &  &  &  &  & Probability-based & Score\\
CTRLEval \cite{ke2022ctrleval} & \checkmark  & & &\checkmark & & & & & & & & & Probability-based & Score\\
\midrule 
UniEval \cite{zhong2022unieval} & \checkmark &  & \checkmark & \checkmark &  &  &  &  &  &  & &   & Generative-based (trained) & Score\\ 
GPTScore \cite{fu2023gptscore} & \checkmark & * & \checkmark & * & * & * & * & \checkmark & * & * &* & *  & Probability-based & Score\\ 

ICE \cite{jain2023multi} &  \checkmark  &  * & \checkmark & \checkmark & * & * & * & * & * & *& * & *  & Probability-based & Score\\ 

Implicit Score \cite{chen2023exploring}&  * & * & * & * & * & * & * & * & * & * & *& *  & Probability-based & Score\\ 

Explicit Score \cite{chen2023exploring}&  * & * & * & * & * & * & * & * & * & * & *& *  & Generative-based & Likert\\ 

COAScore  \cite{gong2023coascore} &  * & * & * & * & * & * & * & * & * & *& * & *  & Generative-based & Likert\\ 

CheckEval \cite{lee2024checkeval} &  * & *  & * & * & * & * & * & * & * & *& * & *  & Generative-based & Score\\ 
DELTASCORE \cite{xie2023deltascore} & \checkmark &  & \checkmark & \checkmark &  &  & \checkmark &  &  & \checkmark & &   & Probability-based & Score \\ 
COHESENTIA \cite{maimon-tsarfaty-2023-cohesentia} & &&&\checkmark& &&&&&&& & Generative-based (trained) & Likert\\
NOV\_COHERENCE \cite{maimon2023novel_coh} & &&&\checkmark& &&&&&&&&  Generative-based (trained) & Likert\\
PERSE \cite{maimon-tsarfaty-2023-cohesentia}& &&&& \checkmark & & & & \checkmark & \checkmark & &  \checkmark & Generative-based (trained) & Likert\\
G-EVAL \cite{liu2023gpteval}  & \checkmark &  *& \checkmark & \checkmark & * & * & * & * & * & \checkmark &*& * & Generative-based & Score \\

FairEval \cite{wang2023faireval} & * & * & * & * & * & * & * & * & * & *& * &* & Generative-based & Comparison/Likert \\
WideDeep \cite{zhang2023wider} & \checkmark & * & * & \checkmark & * & * & * & *& \checkmark & * & * &* & Generative-based & All Formats \\
ChatEval \cite{chan2023chateval} & * & * & * & * & *& * & * & * & * & * & * &* & Generative-based & All Formats \\
SCALEEVAL \cite{chern2024scaleeval} & * & * & * & * & * & * & * & * & * & \checkmark &*& * & Generative-based & Likert \\
TIGERSCORE \cite{jiang2023tigerscore} & \checkmark &  & \checkmark & \checkmark &  &  &  &  &  &  & &  & Generative-based (trained) & Error Analysis \\
AUTO-J \cite{li2023autoj} & \checkmark & * & \checkmark & \checkmark & * & * & * & \checkmark & \checkmark & \checkmark &* & \checkmark & Generative-based (trained) & Likert \\
\bottomrule
\end{tabular}
\end{table}

%% file: sections/5.Traditional.tex
\section{Traditional Evaluation} \label{sec:tradition}
In this section, we discuss traditional evaluation metrics. From Sections \ref{sec:lexical} to \ref{sec:trained}, we analyze metrics for evaluating textual stories, categorized by their method types (Section \ref{sec:method_type}). In Section \ref{sec:multimodal}, we delve into metrics for evaluating multi-modal relevance and the quality of visual output.

\subsection{Lexical-based Metrics} \label{sec:lexical}
As mentioned in Section \ref{sec:method_type}, lexical-based metrics process the story as a bag of tokens or n-grams. The most popular metrics such as {BLEU} \cite{papineni2002bleu}, {METEOR} \cite{banerjee2005meteor}, {ROUGE} \cite{lin2004rouge}, and {CIDEr} \cite{vedantam2015cider} focus on the lexical overlap between candidates and the reference story. These methods can assess the \textit{Overall Quality} of the generated results. 


Some metrics, on the other hand, evaluate a specific aspect of a story: (1) \textit{Fluency} can be measured by calculating intra-story repetition. The {Repetition Percentage} metric \cite{shao2019long} computes the percentage of tokens in the generated story that repeat
at least one 4-gram; the {Intra-story Repetition} metric \cite{yao2019plan} measures the repetition within each generated story through 3-gram word overlaps. 
(2) {TAACO} \cite{crossley2016tool} is a freely available text analysis tool for evaluating \textit{Coherence}. It incorporates over 150 lexical-based indices (such as sentence overlap and paragraph overlap) and outputs a Likert-type scale as the final score. (3) The {Coverage} metric \cite{guan2020knowledge} measures \textit{Commonsense} by counting the average number of knowledge triples that appear in each generated story.

There are also lexical-based metrics focusing on the \textit{Diversity} of stories generated by a model: (1) To access \textit{General Diversity}, for reference-based metrics, the {Backward BLEU} metric \cite{shi2018bbl} measures diversity by applying generated texts as reference, evaluating each test set text with a BLEU score. The {MS-Jaccard} metric \cite{montahaei2019jointly} assesses both quality and diversity, calculating n-gram overlap between generated and referenced stories using the Jaccard index. For reference-free methods, the {Distinct-n} metric \cite{li2015diversity} computes the average ratio of distinct n-grams to total n-grams. The {Self-BLEU} metric \cite{zhu2018self_bleu} computes the average BLEU score of each generated story using all generated stories as reference. The {Inter-Story Repetition} metric \cite{goldfarb2020content} examines 3-gram repetition across stories. 
(2) Specifically, \citet{fan2019strategies} and \citet{goldfarb2020content} access \textit{Event Diversity}, proposing {Unique Verbs} and {Diverse Verbs} metrics that calculate the number and percentage of unique verbs that are not one of the top 5 most frequent verbs in the story. 



\subsection{Embedding-based Metrics} \label{sec:embedding}
Although lexical-based metrics are widely used, they demonstrate a low correlation with human annotations \cite{sulem2018bleuisnot} and show poor semantic comprehension.
Embedding-based metrics \cite{ng2015rouge_we,kusner2015word,clark2019sentence,zhang2019bertscore,zhao2019moverscore,colombo-etal-2021-automatic,depth_score,infolm_aaai2022}, which evaluate based on embeddings from strong pre-trained models, can perform much better in terms of semantic comprehension. In this section, we discuss the metrics that directly use the pre-trained models, without further fine-tuning. Such methods evaluate the equivalence between the embeddings of the target text and references using a matching algorithm, as shown in the model structure on the left side of Figure \ref{fig:eval_model} (a). Generally, the key point is to propose better-designed matching algorithms based on stronger pre-trained models.

Earlier approaches apply the pre-trained word2vec model \cite{mikolov2013efficient}. {ROUGE-WE} \cite{ng2015rouge_we}, the variant of ROUGE, replaces the hard lexical matching in the ROUGE function with soft matching based on the cosine similarity of word2vec embeddings. The matching algorithm of {Word Mover’s Distance (WMD)} \cite{kusner2015word} uses the concept of Earth Mover's Distance \cite{Rubner1998earth}. It calculates the minimum cumulative distance that words in the generated text must travel to reach words in the reference text within the word2vec embedding space. 
However, WMD's reliance on word embeddings limits its ability to capture whole-sentence semantics. To address this, \citet{clark2019sentence} introduce {Sentence Mover's Similarity (SMS)}, which uses sentence embeddings. They also introduce {Sentence and Word Mover's Similarity (S+WMS)}, which uses both word and sentence embeddings. SMS and S+WMS that incorporate sentence-level semantic comprehension would be more appropriate for story evaluation.

More recent methods apply stronger pre-trained models such as BERT \cite{devlin2018bert}. 
BERTScore \cite{zhang2019bertscore}, a popular embedding-based method using contextual BERT embeddings, matches tokens in generated and reference texts via cosine similarity. It can be adapted to assess input relevance by averaging the BERTScore between the source input and each story sentence, termed BERTScore-Source \cite{callan2023interesting}. 
Unlike BERTScore, which applies hard alignments (one-to-one) for words in the generated text and one reference, MoverScore \cite{zhao2019moverscore} provides soft alignments (many-to-one) between the generated result and multiple references. It employs a matching algorithm similar to WMD, leveraging Earth Mover Distance to compute the semantic distance. As the experiments conducted in \citet{chhun2022hanna}, MoverScore shows slightly better performance in story evaluation than BERTScore.

The previous metrics evaluate based on encoded vectors. Some metrics \cite{colombo-etal-2021-automatic,depth_score,infolm_aaai2022}, however, treat the output of pretrained models as probability distributions rather than vector embeddings. They then calculate semantic similarity through these probability distributions. In our taxonomy, we categorize these methods as special embedding-based methods.  {BaryScore} \cite{colombo-etal-2021-automatic} models the output of multiple layers as a probability distribution, aggregates layer information using the Wasserstein barycenter, and computes similarity using WMD's matching function. {DepthScore}  \cite{depth_score} obtains discrete probability distributions of two texts and calculates their similarity by extending univariate quantiles to multivariate spaces. {InfoLM} \cite{infolm_aaai2022} masks each token to derive the discrete probability distribution for each text. It aggregates these distributions using a weighted sum and then calculates the similarity. These metrics \cite{colombo-etal-2021-automatic,depth_score,infolm_aaai2022} show better performance in tasks like Summarization (focusing on faithfulness), but no clear advantage in story evaluation \cite{chhun2022hanna}. Possibly because they do not capture higher-level semantic comprehension.


{TAACO 2.0} \cite{crossley2019tool} is an upgrade from TAACO 1.0 \cite{crossley2016tool}, measuring the aspect of \textit{coherence}. It introduces several new indices tied to local and global cohesion based on semantic embeddings. It also supports reference-based indices that calculate lexical and semantic overlap between a candidate text and a reference text.


\subsection{Probability-based Metrics} \label{sec:probability}
Probability-based methods \cite{Bahl_Jelinek_Mercer_1983,yuan2021bartscore,ke2022ctrleval} count the evaluation score using the generation probability of a story based on pre-trained models, as shown in Figure \ref{fig:eval_model} (b). Such methods are motivated by the idea that better stories will have a high generation probability.

{Perplexity} \cite{Bahl_Jelinek_Mercer_1983} might be the most widely used  probability-based metric, which calculates the negative log-likelihood of a text sequence generated by a language model. A lower perplexity score indicates better quality. Existing works usually apply a powerful pre-trained generative model like GPT-2 \cite{radford2019gpt2}, or further fine-tune it on the story generation training set (Equation \ref{eqn:gen}).

With the proposal of models like BART \cite{lewis2019bart}, which excel in both text generation and comprehension tasks, it's possible to access specific evaluation aspects using probability-based metrics.  {BARTScore} \cite{yuan2021bartscore} calculates the per-token probability of the BART model. By altering the source and the generated text, it can assess various aspects. For example, it can measure relevance and text quality by calculating the generation probability of the story based on the input premise. 
{CTRLEval} \cite{ke2022ctrleval} designs more elaborate text infilling tasks for different evaluation aspects: sentence cohesion (coherence), consistency, and relevance, achieving better performance than BARTScore.


\subsection{Trained Metrics}\label{sec:trained}
As mentioned in Section \ref{sec:method_type},
such metrics require training on the evaluation benchmark to enhance evaluating quality. Some metrics also train on the text generation dataset to improve the model's comprehension ability. In this Section, we first present the commonly used training objects in Section \ref{sec:training_obj}, and then discuss the trained metrics in Section \ref{sec:trained_metrics}.

\subsubsection{Training Objects} \label{sec:training_obj} Given a textual story $y$ constructed with $n$ sequential sentences $\{s_i\}_i^n$ and k tokens $\{w_i\}_i^k$, our goal is to obtain a evaluation result $\mathcal{S}$ achieved by function $f(y)$. For evaluations based on source input $x$ and reference $r$, $\mathcal{S}$ is obtained by $f(y,x,r)$. In the following discussion, we present the loss function for the basic condition. 
For trained metrics, the common training tasks and correlated objectives are as follows:

\paragraph{Story Generation} For probability-based and generative-based methods, the generative model can be trained on the story generation dataset to improve the high-quality text's generation probability. The corresponding negative log-likelihood loss is as follows:
\begin{equation} \label{eqn:gen}
    \scalemath{0.95}{\mathcal{L}_{gen} = - \frac{1}{k} \sum_{t=1}^k log p(y_t|y<t; \theta ),}
\end{equation}
where $ \theta$ are the parameters of the generative model.
\paragraph{Discriminative/Contrastive Learning} Under the evaluation target that high-quality text should receive higher scores than low-quality ones, some methods construct contrastive story pairs ($y^+, y^-$). Given an evaluation dataset, $y^+$ is the text with a higher score than the negative sample $y^-$. Without evaluation datasets, $y^+$ is the ground truth output, and $y^-$ is either randomly selected or carefully crafted. The model is then trained using the discriminative loss:
\begin{equation} \label{eqn:dis}
     \scalemath{0.93}{\mathcal{L}_{dis} = max(0, \delta - f(y^+) + f(y^-) ),}
\end{equation}
where $\delta$ is a threshold parameter. It suggests that the score of a positive sample should exceed the score of a negative sample by at least a margin of $\delta$.
\paragraph{Evaluation Result Classification} If the evaluation result is formatted as a Likert-type scale or boolean output, a classifier model can be trained to predict appropriate classes, usually trained with  the classification loss:
\begin{equation} \label{eqn:cls}
    \scalemath{0.95}{\mathcal{L}_{cls} = - \mathcal{S}_c \cdot logf(y),}
\end{equation}
where $\mathcal{S}_c$ denotes a one-hot encoded vector representing the ground truth class, and $f(y)$ outputs the vector of predicted probabilities for each class.
\paragraph{Evaluation Result Generation}
As shown in Figure \ref{fig:eval_model}, for methods that directly generate a numeric score with the regressor (like BLEURT, using an MLP layer for score generation) can be trained using regression loss as follows:
\begin{equation} \label{eqn:regress}
     \scalemath{0.95}{\mathcal{L}_{regress} = ||\mathcal{S}-f(y)||^2,}
\end{equation}  

where $\mathcal{S}$ is the ground truth score in the evaluation dataset, and $f(y)$ yields the predicted score.

On the other hand, for generative-based methods shown in Figure \ref{fig:eval_model} (c), all evaluation output formats can be converted into text. For example, the Likert-type format can be expressed as integer text, and the boolean format can be represented as ``yes'' or ``no''. So, all evaluation problems can be turned into text generation problems and trained using cross-entropy loss as follows:
\begin{equation} \label{eqn:ce}
    \scalemath{0.85}{ \mathcal{L}_{CE} = - \frac{1}{l} \sum_{t=1}^l logp(\mathcal{ST}_t|\mathcal{ST}<t, y; \theta ),}
\end{equation}
where $\mathcal{ST}$ refers to the text converted from the ground truth evaluation result, containing $l$ words, and $\theta$ represents the parameters of the generative model.

\paragraph{Sentence Reordering}This is a self-supervised task proposed by \citet{lapata2003probabilistic}, which means to reorder shuffled sentences to their original sequence. This task can be applied to access the coherence of a story \cite{maimon2023novel_coh}. There are typically two task settings:
(1) Classification-based \cite{Prabhumoye2020Topological}, which takes a pair of sentences <$s_i,s_j$>($i<j$) and determines whether $s_i$ comes before $s_j$. This is trained using the classification loss.
(2) Generation-based \cite{logeswaran2018sentence_order}, which involves all the sentences and predicts their appropriate order. This is trained using the generation loss.
\paragraph{Instruction Tuning \cite{zhang2023instruction}} This training object is usually applied by LLM-based models (Section \ref{sec:llm_trained}), which fine-tune the LLMs with a large-scale instruction dataset for evaluation. The loss function is similar to Equation \ref{eqn:ce}, with instructions as additional input.

\subsubsection{Trained Metrics} \label{sec:trained_metrics}
 This subsection discusses the trained metrics proposed or can be adopted for story evaluation.
 
{RUBER} \cite{tao2018ruber} is a hybrid metric.
Its reference-free part RUBER$_u$ uses embeddings of the source and target text, and trains the model using contrastive loss (Equation \ref{eqn:dis}). The negative sample is randomly selected from the dataset. {RUBER-BERT} \cite{tao2018ruber_bert} extends the RUBER score by employing BERT contextual embeddings, resulting in improved correlation with human evaluations. {BLEURT} \cite{sellam2020bleurt} aligns more closely with human annotators due to its training setup. It is first pre-trained on large-scale synthetic data, using various automatic metrics (such as BLEU and BERTScore) as supervision signals. It is then fine-tuned on human rating scores. The loss function is shown in Equation \ref{eqn:regress}. The model structure is depicted on the left side of Figure \ref{fig:eval_model}, predicting a score based on the BERT embeddings of the target and reference texts.

The series of COMET scores are built upon the stronger XLM-RoBERTa model \cite{conneau2019xlm_roberta}. Reference-based {COMET} \cite{rei2020comet} combines the embeddings of the target and reference text into one single vector which is then passed to a regression model. Reference-free {COMET-QE} evaluates quality using only the source and target text as input. Both  COMET and COMET-QE are trained on human evaluation data using the regression loss shown in Equation \ref{eqn:regress}.  {COMETKiwi} \cite{rei2022cometkiwi} utilizes the same training method, while adapting the predictor-estimator architecture of OPENKIWI \cite{kepler2019openkiwi} (as shown in the right side of Figure \ref{fig:eval_model} (a)). {COMET22} \cite{rei2022comet} incorporates additional pre-training data and larger encoder models, resulting in enhanced performance. Although the series of COMET scores are proposed for the machine translation task, they also perform well on reference-based story evaluation \cite{jiang2023tigerscore}.

{UNION} and {MANPLTS} are specifically proposed for story evaluation. {UNION} \cite{guan2020union} trains a model to distinguish human-written stories from negative samples. These samples, focusing on \textit{fluency}, \textit{coherence}, and \textit{commonsense}, are created using four techniques: lexical and sentence-level repetition, random keyword and sentence substitution, sentence reordering, and negation alteration.
On the other hand, {MANPLTS} \cite{ghazarian2021MANPLTS} trains a classification model primarily focused on \textit{coherence}. They create negative samples by introducing plot-level incoherence sources. The plot manipulations include logical reordering, contradiction insertion, repetition insertion, and random sentence substitution. These manipulated plots are then fed into generation models to produce implausible stories as negative samples.
\citet{maimon2023novel_coh} propose an evaluation metric, {NOV\_COHERENCE}, to thoroughly assess \textit{coherence}. It is first pre-trained on various tasks that contribute to coherence detection (sentence reordering, irrelevant sentence recognition, etc.) and then trained on a coherence evaluation dataset. Their experimental results demonstrate that these pre-training tasks help the model achieve better performance in coherence evaluation.

\subsection{Multi-modal Tasks Evaluation}\label{sec:multimodal}

When evaluating a multi-modal story, we should measure the cross-modal relevance (Section \ref{sec:multi_modal_rel}) for both visual-to-text and text-to-visual tasks. When evaluating visual-to-text generation, the measurement of generated textual stories remains the same as discussed in previous sections. Whereas for text-to-visual generation tasks, we should evaluate the quality of the visual outputs (Section \ref{sec:visual_quality}), taking into account their quality and visual coherence. 

\subsubsection{\textbf{Multi-modal Relevance}} \label{sec:multi_modal_rel}  {Visual Captioning Accuracy} \cite{maharana2021improving, maharana2021integrating} is commonly used in early research on story visualization. It generates captions for created images and measures the similarity between these captions and the input story. With the introduction of the CLIP model, which effectively aligns text and visuals in the same embedding space, {ClipScore} \cite{hessel2021clipscore} and {EMScore} \cite{emscore} are proposed. These methods evaluate the similarity of text and visual content and can effectively assess the multi-modal relevance of textual story and visual story.

The former metrics separately measure each sentence's relevance to the correlated visual content. To measure the \textit{global relevance} of the whole story and the visual content, {\citet{maharana2021improving}} train a Hierarchical-DAMSM model that extracts global representations for both the story and the image sequence, using their similarity to compute the retrieval-based R-Precision.  {\citet{ning2023album}} apply the earth mover's score \cite{rubner2000earth} to measure the distance between the distribution of the album images and the generated stories in the CLIP embedding space.

Some metrics \cite{wang2022rovist,surikuchi2023groovist} specifically address the\textit{ global multi-modal relevance }in visual storytelling. {RoViST} \cite{wang2022rovist} proposes three evaluation metrics for visual grounding, coherence, and non-redundancy, achieving good overall evaluation performance. However, its RoViST-VG metric, which measures multi-modal relevance, overlooks temporal alignment -- it doesn't consider whether the order of entities in the story matches their order in the image sequence. To address this, \citet{surikuchi2023groovist} conduct an extended analysis of existing metrics and propose a novel metric, {GROOViST}, which is robust to temporal misalignments and correlates with human intuitions about grounding. Specifically, they utilize CLIP \cite{radford2021clip} features to improve visual grounding abilities. To address temporal misalignments, they penalize noun phrases (NPs) that are poorly grounded with temporally relevant images. Although GROOViST performs better in assessing temporal misalignment, simply assigning negative values for poorly grounded NPs isn't entirely reasonable. These NPs might represent a review of previously appeared visual content or could be logically correlated with current visual components. Further exploration is needed.


\subsubsection{\textbf{Visual Output Evaluation}} \label{sec:visual_quality}
 {FID} \cite{heusel2017gans} and {FVD} \cite{unterthiner2019fvd} are commonly used to assess the \textit{quality} of visual outputs. They measure the similarity between the features of the visual content and those of real-world images and videos. These features can be extracted by any model, such as standard features from Inception-V3 (image) \cite{szegedy2016rethinking} and I3D (video) \cite{carreira2017quo}, or from more powerful models like CLIP (image) \cite{radford2021clip} and InternVideo-MM-L-14 (video) \cite{wang2022internvideo}. {DOVER} \cite{wu2022disentangling} can be used to measure the \textit{perceptual quality} of visual outputs. It predicts the average human subjective perception of a video.

Besides the quality of the output visual stories, the \textit{coherence} of the visual scenes is also important. When evaluating the consistency of a generated video or an image sequence, two factors are typically considered: character and background consistency \cite{rahman2023make,maharana2021integrating}, usually measured by the accuracy and F1-Score.

In tasks of continuous visual storytelling \cite{maharana2022storydall,bugliarello2024storybench}, the generated video or images are constrained by the visual prompt, leading to less randomness. For example, a car wouldn't have a random visual appearance. Therefore, we can directly compare these visual contents with the ground truth by calculating the cosine similarity of their normalized visual features. 

%% file: sections/6.LLM_related.tex
\section{LLM-Based Evaluation} \label{sec:llm}
The development of large language models (LLMs) has led to significant advancements in automatic comprehension and generation. This also encourages several LLM-based evaluation methods \cite{gao2024llm-based, li2024leveraging}. As verified in \citet{wang2023chatgpt}, LLM has a much higher correlation with humans than traditional metrics. Additionally, it is capable of providing a human-like reasoning process, demonstrating much stronger reliability and interpretability.  We discuss existing LLM-based evaluation methods from Section \ref{sec:llm_embedding} to \ref{sec:llm_trained}, along with their pros and cons in Section \ref{sec:llm_take}.

\subsection{Embedding-based Metrics} \label{sec:llm_embedding} Similar to the traditional embedding-based methods mentioned in Section \ref{sec:embedding}, researchers can utilize embeddings calculated by more advanced LLMs to achieve better performance \cite{es2023ragas}. Clearly, these methods have limitations -- they lack explainability and struggle to encompass various evaluation aspects.

\subsection{Probability-based Metrics} \label{sec:llm_pro} Such methods \cite{chen2023exploring,fu2023gptscore,xie2023deltascore} count the evaluation score using the generation probability of LLMs. By designing different inputs and outputs, these probability-based metrics can address various aspects. The {Implicit Score} \cite{chen2023exploring} simply forms the evaluation (overall quality or specific aspect) as a binary Yes or No question, and calculates based on the generation probability of answering ``yes''.

{GPTScore} \cite{fu2023gptscore} calculates the generation probability of the target story, based on the source input, task specification, and criteria definition. Since the probability of a sentence can vary due to superficial differences such as word order and sentence structure, such probability-based methods are susceptible to likelihood bias, particularly in aspects like relevance. \citet{ohi2024likelihood} quantify and explore the impact of this bias. Additionally, they propose a {Likelihood-Bias-Mitigation} method to mitigate likelihood bias by using highly biased instances as few-shot examples for in-context learning.

Inspired by the idea that higher quality stories are more affected by perturbation than lower quality ones (for example, introducing typos could impact a fluent story more than a non-fluent one), \citet{xie2023deltascore} propose a novel method named {DELTASCORE}. This method evaluates the likelihood difference between stories before and after applying perturbation strategies related to specific aspects.

\subsection{Generative-based Metrics} \label{sec:llm_gen} 
Such methods can also be defined as prompt-based methods, which attempt to prompt strong LLMs to automatically evaluate the results. In other words, they let LLMs serve as human annotators. One of the biggest advantages of generative-based metrics is interpretability, by generating the reasoning process for the evaluation result. The key challenge of these methods is to design proper prompts and effective frameworks.

Aggregating evaluations through more detailed and easier sub-tasks can lead to better performance than directly assessing the overall quality. \citet{saha2023branch} propose the {BRANCH-SOLVE-MERGE (BSM)} method, which breaks down the overall evaluation into several sub-tasks. LLM generates results for each sub-evaluation and then aggregates them. \citet{zhang2023wider} propose {WideDeep}, a multi-layer LLM evaluator. In its first layer, each neuron handles a specific task of quality evaluation. In higher layers, each neuron integrates and abstracts the previously learned local evaluation information to generate a more comprehensive evaluation result. This multi-layer comprehension leads to improved performance. 
COAScore \cite{gong2023coascore} prompts the LLM to generate a chain of aspects (such as those shown in Section \ref{sec:criteria}) for evaluation. It scores each generated aspect and uses the chain-of-aspect knowledge (definitions and scores) to achieve the final evaluation result. {CheckEval} \cite{lee2024checkeval} breaks down each evaluation aspect into more detailed sub-aspects, develops a checklist for each dimension, and achieves better performance.

LLMs can further enhance evaluation performance, consistency, and robustness by merging or debating on multiple evaluation results. 
{FairEval} \cite{wang2023faireval} generates multiple pieces of evidence before assigning final ratings. G-Eval \cite{liu2023gpteval} calculates the final score as a probability-weighted summation of different Likert-type scales. In {ChatEval} \cite{chan2023chateval}, unique personas are assigned to an LLM, leading to multiple agents engaging in the debate process. These agents can reach a conclusion after one-by-one or simultaneous debates. An additional summarizer can summarize each iteration of a simultaneous debate to add high-level messages. Through multi-agent discussion, more reliable and robust evaluation is achieved. {MATEval} \cite{li2024mateval} further incorporates the Chain-of-Thought \cite{wei2022chain} and self-reflection \cite{madaan2024self} strategies into the multi-agent discussion, improving the performance in open-ended story evaluation. As different LLMs naturally have differences and can comment on the results of others, this omits the process of persona assignment. {\citet{bai2024benchmarking}} propose a peer-examination method that lets LLMs evaluate the results of other LLMs and combine the ranking results.
{SCALEEVAL} \cite{chern2024scaleeval} proposes an agent-debate-assisted meta-evaluation framework. It allows different LLM agents to engage in multi-round debates, resulting in more reasonable evaluation results that correlate highly with human annotators. It also provides the process of discussions to assist humans. SCALEEVAL supports any new user-defined scenarios and criteria, aiding automatic evaluation and new benchmark annotation.

{PORTIA} \cite{li2023split} splits the long answers into multiple segments, aligns similar content, and then merges them back into a single prompt for evaluation. Its main idea is to divide the comparison of long contexts into several comparisons between similar shorter segments, making the evaluation easier and reducing the problem of position bias. This might also be applied to story evaluation, for instance, dividing the evaluation into comparisons of the opening, progression, and ending.


\subsection{Trained Metrics} \label{sec:llm_trained}
Due to the high cost and potential irreproducibility of generative-based metrics, some methods focus on fine-tuning open-source LLMs to create expert models for evaluation. 

Most of them access general evaluation tasks (with creative writing as a sub-domain), including {PandaLM} \cite{wang2023pandalm}, {Prometheus} \cite{kim2023prometheus,kim2024prometheus2opensource}, {Shepherd} \cite{wang2023shepherd}, {Auto-J} \cite{li2023autoj}, {CritiqueLLM} \cite{ke2023critiquellm}, {JudgeLM} \cite{zhu2023judgelm} and {TIGERScore} \cite{jiang2023tigerscore}. They collect large-scale evaluation benchmarks and fine-tune pre-trained LLMs through instruction tuning \cite{zhang2023instruction}. More details of their instruction datasets are summarized in \citet{gao2024llm-based}. Among these trained models, Auto-J \cite{li2023autoj} proposes several referable criteria for each task, while TIGERScore \cite{jiang2023tigerscore} provides detailed error annotations of various aspects. They demonstrate the potential to handle unseen tasks when instructions are built using predefined criteria.

{PERSE} \cite{wang2023perse} and {COHESENTIA} \cite{maimon-tsarfaty-2023-cohesentia} specifically focus on story evaluation. The former emphasizes personalized story evaluation, while the latter concentrates on coherence evaluation.



\subsection{Takeaways} \label{sec:llm_take}
Although LLM-based evaluations achieve much better results than traditional methods, there is still a long way to go in realizing reliable and robust evaluation. Here we would like to summarize some useful strategies and existing limitations.

\paragraph{Helpful Strategies} \label{sec:llm_strategy}
We summarize the generally useful strategies in LLM-based evaluation: (1) In the evaluation prompt, clear and concise instructions are better than complex ones \cite{kim2023better}. (2) Generating the reasoning process or detailed error analysis can aid the final evaluation \cite{kim2023better,chiang2023closer,li2023autoj,jiang2023tigerscore}. (3) Decomposing a complicated task into simpler, clearer sub-tasks is helpful \cite{saha2023branch,zhang2023wider,gong2023coascore,li2023split}. (4) Aggregating multiple evaluations \cite{wang2023faireval}, incorporating a debating process \cite{chan2023chateval,li2024mateval,chern2024scaleeval}, and using multi-agent assessment \cite{chan2023chateval} can all improve the performance and robustness of the results. (5) In-context learning could be helpful \cite{jain2023multi}, especially for personalized evaluation \cite{wang2023perse}.


\paragraph{Limitations} There are four types of discovered biases: (1) Format Bias, which means the optimal performance is only achieved under specific prompt formats. Designing more professional prompts or fine-tuning the model with more diverse instructions can mitigate this problem, but it still remains to be solved. (2) Position bias, a common problem in LLM-based evaluations, which means that LLM exhibits a preference for the first displayed candidate \cite{wang2023faireval}. To address this problem, a simple solution is to combine evaluations using different sample orderings. Specifically for generative-based methods, \citet{wang2023faireval} aggregates results across various orders to determine the final score; for trained methods, \citet{li2023autoj} change the order of the candidates within each training sample to double the training data. (3) Knowledge bias, which means LLM-based evaluators tend to favor the results they have seen or results that are generated by themselves \cite{liu2023gpteval,liu2023evaluate}. This can be partially mitigated by replacing proper nouns, but it remains a very challenging issue. (4) Likelihood Bias, which is showcased in probability-based methods, because the probability of a sentence can vary due to superficial differences, especially in aspects like relevance \cite{ohi2024likelihood}.

There are also concerns about the substantial expense and non-reproducibility of methods based on commercial LLMs. This calls for more research based on open-source models. Proposing evaluation benchmarks for developing stronger evaluation expert models is also valuable.

%% file: sections/7.Discussions.tex
\section{Evaluating Metrics on Story Evaluation Benchmark} \label{sec:evaluate_metric}
In this section, we present the evaluation performance\footnote{The correlation of these metrics with the human-annotated evaluation dataset can be calculated using Pearson \cite{pearson1895vii}, Spearman \cite{spearman1961proof}, or Kendall-Tau \cite{kendall1938new} correlations.} of existing automatic metrics on the story evaluation benchmark, from collected experimental results and our additional experiments. We focus on the most commonly used metrics and those demonstrating exceptional performance. We discuss their advantages and disadvantages, and suggest potential uses for them.

\subsection{Evaluation of Overall Quality}
\begin{figure}[t]
    \centering
    \includegraphics[width=\linewidth]{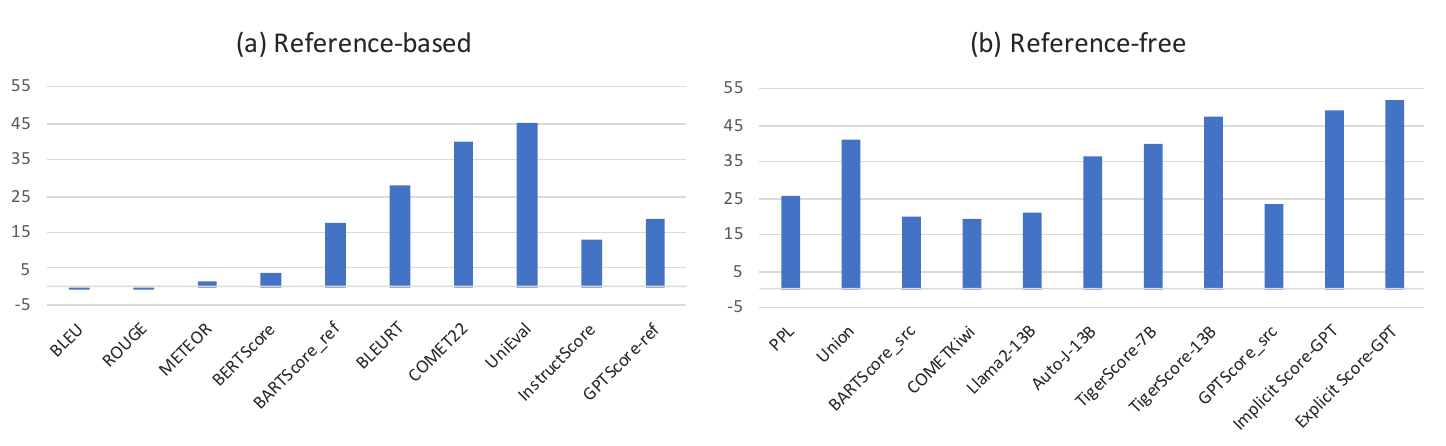}
    \caption{The Pearson Correlation between various metrics and human ratings on OpenMEVA (ROC) benchmark dataset.}
    \label{fig:correlation}
\end{figure}

The evaluation results of automatic metrics on the story evaluation benchmark dataset OpenMEVA\footnote{To the best of our knowledge, it is the story evaluation benchmark that has the highest citation.} are displayed in Figure \ref{fig:correlation}. The representative evaluation metrics involve: (a) \textbf{Reference-based Metrics}, including widely used lexical-based metrics (BLEU \cite{papineni2002bleu}, ROUGE \cite{lin2004rouge}, METEOR \cite{lin2004rouge}), embedding-based and probability-based metrics (BERTScore \cite{zhang2019bertscore}, BARTScore\_ref \cite{yuan2021bartscore}), trained metrics (BLEURT \cite{sellam2020bleurt}, COMET22 \cite{rei2022comet}, UniEval \cite{zhong2022unieval}), and LLM-based scores (InstructScore \cite{xu2023instructscore}, GPTScore\_ref \cite{fu2023gptscore}).
(b) \textbf{Reference-free metrics}, including commonly used probability-based metrics (PPL \cite{Bahl_Jelinek_Mercer_1983}, BARTScore\_src \cite{yuan2021bartscore}), trained metrics (Union \cite{guan2020union}, COMETKiwi \cite{rei2022cometkiwi}), and strong LLM-based scores (Llama2-13B \cite{touvron2023llama2openfoundation}, GPTScore\_src \cite{fu2023gptscore}, Auto-J \cite{li2023autoj}, TigerScore \cite{jiang2023tigerscore}, Implicit/Explicit Score \cite{chen2023exploring}).

Our observations and discussions, which could also be applied to other creative generation tasks, are as follows:
\begin{itemize}
    \item While accessing the overall performance of a textual story, although we have stated that there are no standard answers for creative story generation, reference-based evaluation results remain significant. Metrics such as COMET22 and UniEval, which have been trained on human evaluation benchmarks, can show good correlations with human judgments. It's advantageous to present both reference-based and reference-free metrics. However, for longer stories, reference-based metrics may be more suitable for plot-level matching rather than evaluating the entire story.
    \item Regarding reference-based methods, widely used lexical-based metrics such as BLEU and ROUGE show very low correlation with human evaluation (Figure \ref{fig:correlation}). Embedding-based and probability-based metrics demonstrate stronger performance due to their improved semantic comprehension. However, to better align with human judgment, training on human evaluation benchmarks is necessary. This is verified by the performance of UniEval and COMET22, which show even stronger results than some reference-based LLM metrics.
    \item Regarding reference-free methods, although the metrics based on powerful ChatGPT achieve high performance, they suffer from non-reproducibility and high cost. 
    Open-source metrics that show comparable results are preferable alternatives. Specifically, metrics trained for story evaluation such as Union \cite{guan2020union}, or evaluation expertise LLMs like TigerScore \cite{jiang2023tigerscore} and AUTO-J \cite{li2023autoj} are worth considering. We can further enhance performance by applying additional strategies (Section \ref{sec:llm_strategy}) such as the debating process.
    \item Existing automatic evaluations, even the effective LLM-based methods, have their limitations. Thus human evaluation and collaborative evaluation (Section \ref{sec:collaborative}) still have significant value.
\end{itemize}

\subsection{Evaluation of Sub-Aspects}
\begin{figure}[t]
    \centering
    \includegraphics[width=0.85\linewidth]{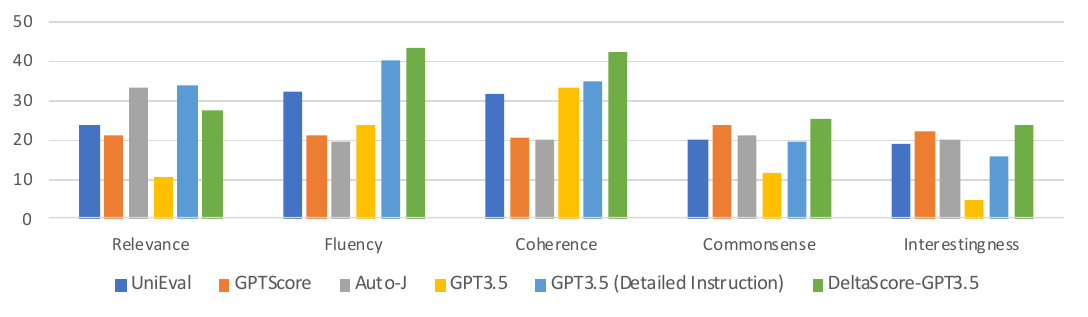}
    \caption{The Kendall Correlation between powerful metrics and multi-aspect human ratings proposed by \citet{xie2023next}. }
    \label{fig:aspects}
\end{figure}

Regarding the evaluation of various aspects, our discoveries and discussions are as follows:
\begin{itemize}
    \item We select powerful metrics capable of evaluating specific aspects and present their performance on the benchmark proposed by \citet{xie2023next}, a dataset covering commonly explored aspects. As shown in Figure \ref{fig:aspects}, evaluation expert models like UniEval and AUTO-J perform comparably or better than  ChatGPT \footnote{We use OpenAI API with the GPT-3.5-turbo model}. Detailed instructions (including criteria definitions and scoring standards) \cite{Chhun2024do} or fine-grained evaluations (such as DeltaScore \cite{xie2023deltascore}) can improve the evaluation of specific aspects. Note that this analysis may require validation on more robust benchmarks. 
    \item According to the experimental results shown in Figure \ref{fig:aspects}, as well as existing research works \cite{chhun2022hanna,Chhun2024do,xie2023deltascore}, LLM-based methods are currently the best proxy for human evaluation of stories. However, they still face challenges in aspects like interestingness, which require further exploration in both general and personalized evaluation \cite{wang2023perse}.
    \item Especially regarding coherence, existing research such as COHENSENTIA \cite{maimon-tsarfaty-2023-cohesentia} and NOV\_COHERENCE \cite{maimon2023novel_coh} can be applied. However, these approaches are not that effective for evaluating long stories. 
\end{itemize}

%% file: sections/8.Collaborative.tex
\section{Collaborative Story Evaluation and Generation} \label{sec:collaborative}

This section discusses the research on human-AI collaborative evaluation and the evaluating considerations for collaborative writing frameworks.
\paragraph{Collaborative Evaluation}Both human evaluation and automatic evaluation have their own advantages and limitations. Human evaluation is considered the gold standard, but it can be time-consuming, expensive, subjective, and inconsistent \cite{clark2021all}. Automatic evaluations, particularly those based on open-source LLMs, can be cost-effective and produce fewer outlier values. However, they still need improvement to better correlate with human standards. Recent studies have explored to combine these advantages through collaborative evaluation. \citet{li2023collaborative} introduce {COEVAL}, a two-stage approach. First, it generates a checklist of task-specific criteria, then conducts instance evaluation. Both stages begin with LLM-generated results, which are subsequently scrutinized by humans. Through this human scrutiny, COEVAL revises approximately 20\% of evaluation scores for ultimate reliability. \citet{kim2023evallm} propose EvalLM, an interactive platform that provides automatic, detailed evaluation outputs based on user-defined criteria. This enables users to efficiently refine task prompts and evaluation criteria. In sum, the main issue lies in designing effective interactive strategies for efficient collaboration between humans and AI systems.

\paragraph{Evaluation of Collaborative Story Generation} 
Today's Large Language Models, particularly commercial ones like GPT-4 \cite{achiam2023gpt} and Claude \cite{wu2023claude}, can match or slightly surpass human writers in most areas \cite{xie2023can, xie2023next}. However, they may still fall short in aspects such as creativity and interestingness \cite{gomez2023confederacy,xie2023next,lee2022coauthor}. Nevertheless, this advancement enhances the feasibility of collaborative story generation -- a process where a person works with model outputs to jointly create a story \cite{ippolito2022creative,branch2021collaborative,clark2018creative}.
Evaluating collaborative writing requires consideration of both \textit{story quality} and \textit{user experience}. Quantitative evaluations of human experience include edit distance \cite{dhillon2024shaping}, percentage of accepted suggestions or applied model-generated stories, generation productivity (words written per unit time) \cite{dhillon2024shaping}, and story completion time \cite{swanson2009say}. \citet{lee2022coauthor} apply the concepts from \citet{storch2002patterns}, redefining equality as the even distribution of writing events between humans and AI, and mutuality as the proportion of user-system interactions among all operations.

User feedback might be more important for evaluating collaborative writing systems. Several studies \cite{li2024value, yeh2024ghostwriter, mirowski2023co, yuan2022wordcraft, goldfarb2019plan, stevenson2022putting} have developed questionnaires to gather this feedback. These questions either assess the overall performance (whether the framework is a good platform and if they would use it again); or assess multiple aspects, focusing on the helpfulness, user-friendliness, user experience, and effectiveness of the collaborative writing framework. Detailed questions can be located in each paper, where Table \ref{table:co_aspects} shows the example questions from \citet{mirowski2023co}. Future research may adapt these questions to align with the specific goals of proposed systems.

\begin{table}[t]
\caption{The common aspects used for evaluating collaborative writing.}
\label{table:co_aspects} 
\fontsize{6.8}{6.9}\selectfont
\begin{tabular}{ll}
\toprule
\multicolumn{1}{c}{\textbf{Aspect}} & \multicolumn{1}{c}{\textbf{Definition} } \\
\midrule
\textbf{Helpfulness} & I found the AI system helpful. \\
\midrule
\textbf{Collaboration} & I felt like I was collaborating with the AI system. \\
\midrule
\textbf{Ease} &  I found it easy to write with the AI system.\\
\midrule
\textbf{Enjoyment} & I enjoyed writing with the AI system. \\
\midrule
\textbf{Expression} &  I was able to express my creative goals while writing with the AI system.\\
\midrule
\textbf{Unique} &  The script(s) written with the AI system feel unique. \\
\midrule
\textbf{Ownership} & I feel I have ownership over the created script(s). \\
\midrule
\textbf{Surprise} & I was surprised by the responses from the AI system. \\
\midrule
\textbf{Proud} & I'm proud of the final outputs. \\
\bottomrule
\end{tabular}
\end{table}

%% file: sections/9.Future_work.tex
\section{Recommendations for Future Explorations} \label{sec:future}
Based on the above survey, we would like to make following recommendations for future research explorations.
\paragraph{Standardized Evaluation Criteria} Although we have analyzed the evaluation criteria in previous research, as shown in Section \ref{sec:criteria}, we hope to encourage further exploration of standardized evaluation criteria, especially for subjective aspects like interestingness that are difficult to evaluate automatically. Specifically for LLM-based evaluation, it's valuable to explore suitable criteria for various LLMs, as different models may favor distinct definitions. For example, we can iteratively refine initial criteria through human feedback \cite{kim2023evallm} or system self-improvement \cite{liu2023calibrating}, based on issues identified in evaluation results using the current criteria.

\paragraph{Story Evaluation Benchmark} Most existing story evaluation benchmarks (Section \ref{sec:benchmark}) are limited to stories generated from prompts in ROCStories \cite{mostafazadeh2016rocstories} and WritingPrompts \cite{fan2018wpdataset}. Future benchmarks should encompass more diverse domains, incorporate more complex aspects, and include longer stories. Additionally, since human evaluation is expensive and time-consuming, it is usually conducted on a subset of a large-scale dataset. Therefore, exploring methods for effective subset sampling is also valuable \cite{ruan2024better}.

\paragraph{Long Story Evaluation} With the development of Large Language Models, it is more possible for automatic evaluation and generation of long story \cite{you2023eipe,lee2022coauthor,Yang2023DOCIL,Zhou2023RecurrentGPTIG}. However, evaluating long narratives presents numerous challenges. Firstly, obtaining human annotations for lengthy stories is difficult. Secondly, encoding and comprehending long contexts presents a challenging task. While existing works have made progress in 0-10K story processing, the automatic generation and evaluation of human-like long stories, such as a Harry Potter fanfic fiction of at least 40K words \cite{mikhaylovskiy2023long}, is still challenging.

\paragraph{Multi-modal Story Evaluation} As discussed in Section \ref{sec:visual2text}, in multi-modal story evaluation, we should consider not only spatial relevance but also temporal relevance and more complex logical relevance. Although some efforts \cite{wang2022rovist,surikuchi2023groovist} have been made to access these challenges, this is still a under-explored domain. 
\paragraph{Personalized Evaluation} Human evaluation might suffer from low inter-annotator agreement, particularly for subjective aspects like interestingness and character development \cite{callan2023interesting}. It's worthwhile to explore personalized evaluation, which is valuable for applications like recommendation systems. \citet{wang2023perse} explore to address the problem of personalized evaluation. They propose two personalized story evaluation datasets, fine-tuning open-source LLM with instructions and few-shot reviewer preference. Further research into more benchmarks and personalized evaluation methods is encouraged.

\paragraph{Fairness Improvement} Fairness improvement (or debiasing) is a crucial issue in LLM-based evaluation. Common problems include position bias, where LLMs favor the first option in a comparison pair; format bias, which refers to their sensitivity to prompts; and knowledge bias, which implies a preference for memorized or generated stories. Despite efforts to mitigate these biases, further research is needed. 

\paragraph{Robustness Analysis} \citet{he2022blind} has developed various stress tests to examine the robustness of metrics derived from medium-size pre-trained language models. They identify several blind spots in metrics such as BERTScore \cite{zhang2019bertscore} and UniEval \cite{zhong2022unieval}. Although newer metrics based on more powerful LLMs show increased robustness, assessing the reliability of these automatic metrics remains essential.

\paragraph{Reliability Exploration} Improving reliability helps ensure people trust and use automatic evaluation results. To address this, existing metrics primarily provide a more human-like rational process \cite{kim2023better} or conduct more reliable error analysis \cite{jiang2023tigerscore}. Future works could also explore ways to mitigate overconfidence and reduce inconsistencies.

\paragraph{Explore the Difference between Human-written and AI-generated Stories} Here we have two objectives: first, to explore the gap between AI-generated and human-written stories in order to enhance the quality of generated content. Second, to develop methods for distinguishing between AI-generated and human-written stories, with the aim of preventing potential harm, such as the spread of fake news stories.

\paragraph{Collaborative Evaluation} Collaborative evaluation can leverage the effectiveness of both human evaluation and automatic evaluation. \citet{callan2023interesting} found that automatic metrics can perform at near-human levels, except for aspects like interestingness. As such, it is reasonable to conduct an automatic evaluation first, providing the result and reasoning process. Based on these outputs, humans can then focus on assessing the aspects that machines seem to struggle with.

%% file: sections/10.Conclusion.tex
\section{Conclusion}
In this survey, we summarize and discuss the evaluation of human-written or automatically generated stories. We first progress with various types of story generation tasks, the story evaluation criteria, and benchmark datasets. We then introduce a taxonomy to categorize existing metrics that are proposed or can be adopted for story evaluation, discussing them in detail. Additionally, we carry out experiments, report their quantitative performance on story evaluation, and provide recommendations. Finally, we propose future directions for story evaluation, which are suitable for general evaluations as well. We hope our survey will help readers understand the developments in story generation and automatic evaluations, while inspiring future research directions.